%% file: iclr2026_conference.tex
\documentclass{article} 
\usepackage{iclr2026_conference,times}

\input{math_commands.tex}

\usepackage{graphicx}
\usepackage{siunitx}
\usepackage{booktabs}
\usepackage{hyperref}
\usepackage{multirow}
\usepackage{url}
\usepackage{subcaption}
\usepackage{wrapfig}
\usepackage[T1]{fontenc}
\usepackage{tabularx} 

\title{\textsc{LFQA-E}: Carefully Benchmarking Long-form QA Evaluation}

\author{Yuchen Fan$^{1,2,*}$, Chen Ling$^{5,*}$, Xin Zhong$^{6,*}$, Shuo Zhang$^{4}$, Heng Zhou$^{2}$, Yuchen Zhang$^{2}$,\\ \textbf{Mingyu Liang$^{4}$}, \textbf{Chengxing Xie$^{2}$}, \textbf{Ermo Hua}$^{3}$,
\textbf{Gang Chen}$^{3}$, \textbf{Zhizhou He}$^{3}$, \textbf{Cheng Huang}$^{3}$, \\
\textbf{Ning Ding}$^{3,\dagger}$, \textbf{Bowen Zhou}$^{2,3,\dagger}$ \\
$^1$ Shanghai Jiao Tong University \quad
$^2$ Shanghai AI Lab \quad $^3$ Tsimnghua University \quad \\
$^4$ Beijing University of Posts and Telecommunications \quad
$^5$ Zhejiang University \quad  \\
$^6$ University of Tokyo \quad \\
$^*$Equal contributions~~ $^\dagger$Corresponding author \vspace{1mm} \\
\texttt{yuchenfan48@gmail.com} \quad \texttt{zhoubowen@tsinghua.edu.cn}
\\
\\
}

\begin{document}

\maketitle

\input{main/abstract}
\input{main/introduction}
\input{main/related_work}
\input{main/methodology}

\input{main/experiments}
\input{main/analysis}

\input{main/conclusion}


\section*{Ethics Statement}
This paper proposes a benchmark consisting of questions and references from examination papers and online forums. We perform data filtering to ensure that there is no offense in the benchmark data.

\section*{Reproducibility Statement}
We have provide our settings comprehensively for data construction and evaluation so that the result can be reproduced using the same settings and prompt.

\section*{Acknowledgement}
We thank the China Construction Third Engineering Bureau Group Co.,Ltd. for their support.
w
\bibliography{iclr2026_conference}
\bibliographystyle{iclr2026_conference}

\newpage
\appendix
\input{main/appendix}

\end{document}

%% file: math_commands.tex
\usepackage{amsmath,amsfonts,bm}

\def\eqref#1{equation~\ref{#1}}

\def\1{\bm{1}}

\DeclareMathAlphabet{\mathsfit}{\encodingdefault}{\sfdefault}{m}{sl}
\SetMathAlphabet{\mathsfit}{bold}{\encodingdefault}{\sfdefault}{bx}{n}



%% file: main/abstract.tex
\begin{abstract}
Long-Form Question Answering (LFQA) involves generating comprehensive, paragraph-level responses to open-ended questions, which poses a significant challenge for evaluation due to the richness of information and flexible response format. Existing LFQA-evaluation benchmarks often lack reference answers and are limited in size and topic coverage, reducing their reliability. To address this gap, we introduce \textsc{LFQA-E}, a well-constructed, multilingual, and reference-based benchmark designed to rigorously evaluate automatic metrics for LFQA. \textsc{LFQA-E} comprises 1618 questions and 7323 pairwise comparisons across 15 topics, drawn from diverse sources such as online queries and examination questions, thereby enabling a comprehensive assessment of evaluation metrics. We examine five categories of metrics, encompassing 17 specific methods, using \textsc{LFQA-E}. The results demonstrate that none of the existing automatic metrics perform comparably to human judgments, highlighting their inability to capture the dense information in long-form responses. Furthermore, we present a detailed analysis of the failure cases and the generalization capacity of these metrics, offering insights to guide the future development of LFQA evaluation methods. The benchmark and code are available at \url{https://github.com/YuchenFan48/LFQA-E}.
\end{abstract}

%% file: main/introduction.tex
\section{Introduction}

Long-form Question Answering (LFQA) \citep{fan2019eli5longformquestion} 
targets at generating in-depth, paragraph-level responses to open-ended questions. It requires models to have comprehensive domain-specific knowledge or use evidence from retrieved documents
\citep{nakano2022webgptbrowserassistedquestionansweringhuman, akash2023longformquestionansweringiterative} to provide relevant and accuracy answers. 
Despite efforts to enhance the quality of long-form answers, 
developing automatic and reliable evaluation metrics for LFQA is still underexplored.

Evaluating long-form answers presents significant challenges, as evaluators must possess comprehensive domain knowledge. Previous manual evaluations typically employed crowd-sourced workers for annotation. However, their limited domain expertise inevitably compromises reliability. In contrast, expert annotation would ensure higher quality, while the cost of employing experts to annotate large-scale datasets is prohibitive. Consequently, automatic evaluation metrics are essential. In automatic evaluation of LFQA, ROUGE \citep{lin-2004-rouge} has been widely adopted. However, \citet{krishna2021hurdlesprogresslongformquestion} argue that ROUGE provides limited informativeness in long-form contexts, weakening its reliability. With the advancement of LLMs \citep{openai2023chatgpt, openai2024hello} and Large Reasoning Models (LRMs) \citep{deepseekai2025deepseekr1incentivizingreasoningcapability,zhang2025survey}, numerous studies have leveraged these models to develop evaluation metrics through various approaches \citep{chang2023surveyevaluationlargelanguage}, including prompting \citep{wei2023chainofthoughtpromptingelicitsreasoning,zhou2025reso}, fine-tuning \citep{li2023generativejudgeevaluatingalignment,fan2024evaluating,jiang2024tigerscorebuildingexplainablemetric}, and training LLMs as Reward Models (RMs) \citep{liu2024skywork, chen2025rmr1rewardmodelingreasoning}, either generative or scalar-based. Despite the advances of evaluation metrics, determining which metrics are most effective and best aligned with human judgment for LFQA evaluation requires systematic verification and benchmarking.

Previous benchmark for LFQA evaluation \citep{xu2023criticalevaluationevaluationslongform} samples records from reddit/ELI5, hiring experts to annotate the better one between two responses without references, and test alignment between automatic evaluation metrics and expert labels. However, their benchmark has several limitations: \textbf{1) Lack of authorized references} A reference answer provides a baseline for assessing whether a response covers key details and maintains factual accuracy. Without ground-truth references, the comparison between metrics may be unfair, and evaluations without clear criteria or rubrics are inherently unreliable. \textbf{2) Limited diversity} The benchmark consists of only $\textit{260}$ examples, all in English, constraining its linguistic and topical diversity. Moreover, it treats the comparison as an A/B task, but in real scenarios, a ``tie" option always exists.
\begin{figure*}[t]
    \centering
    \includegraphics[width=0.98\linewidth]{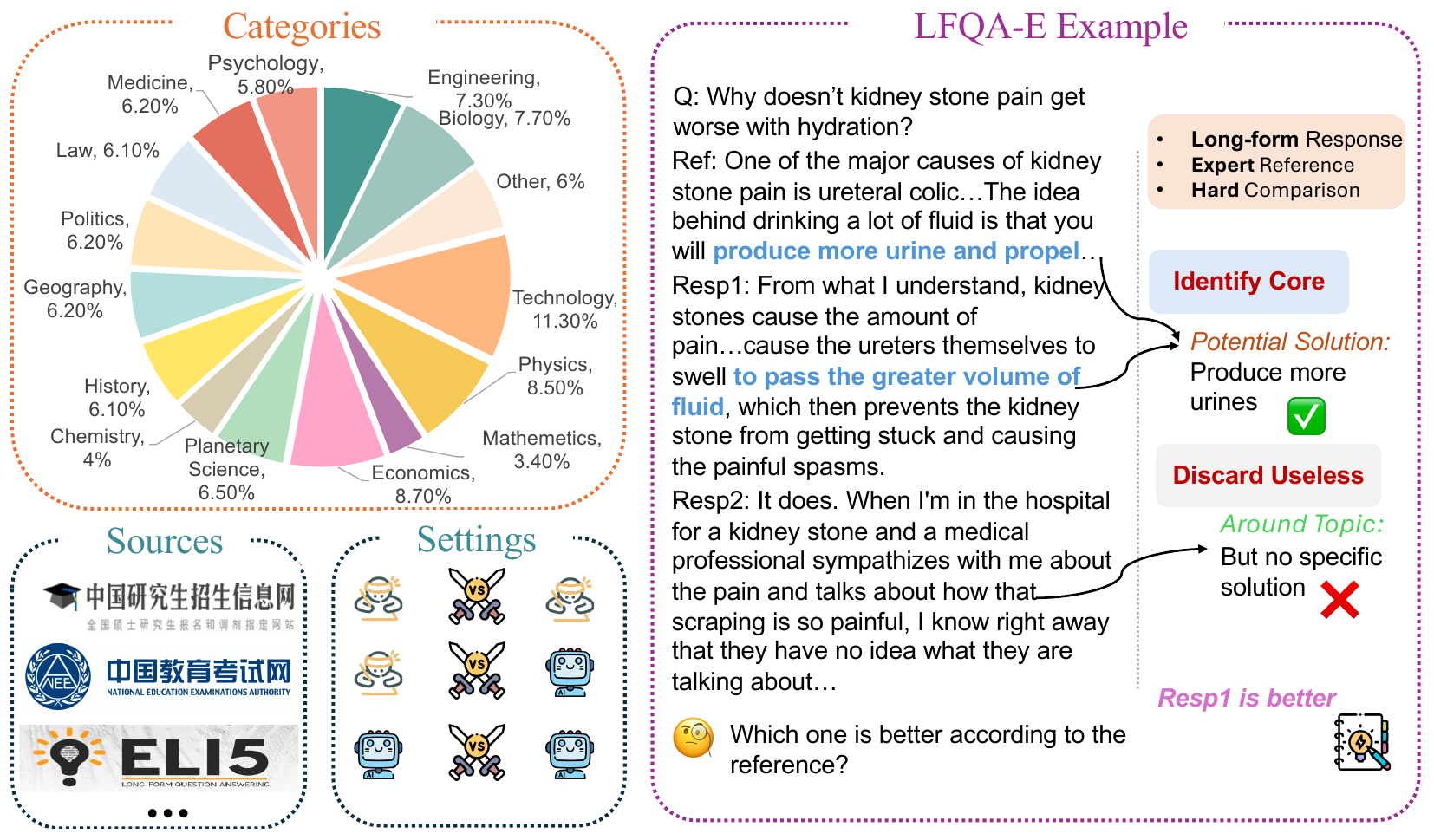}
    \caption{The figure shows the overview of \textsc{LFQA-E}. The left side displays the categories, sources, and three settings, showcasing its diversity. The right side illustrates an example of \textsc{LFQA-E}.}
    \label{fig:overview}
    \vspace{-8mm}
\end{figure*}

To fill the gap, we introduce \textsc{LFQA-E}, towards evaluating the ability of different metrics. 1) To evaluate whether current automatic evaluation metrics can select a better one from two nuanced responses, we gather references that are examined by the experts, and judge based on them. To ensure the difficulty, we choose human responses based on their upvotes or their scores, and model responses based on two models with comparable capabilities. 2) To analyze the systematic differences in validity among evaluation metrics, we rigorously assess the performance from several aspects. First we evaluate them based on three settings, i.e, human vs human (\textit{h v. h}), human vs model (\textit{h v. m}), and model vs model (\textit{m v. m}). Moreover, we collect multilingual responses, i.e, English and Chinese, and multiple domain-specific responses, e.g., Engineering, Law, Medicine, to ensure a thorough analysis. 3) To prevent data contamination, we collect data from offline examination, i.e., College Entrance Examination Simulation Questions (CEESQ) and Postgraduate
Entrance Examination Questions (PEEQ) and online platform questions (reddit/ELI5) from the recent half-year. The overview of \textsc{LFQA-E} Benchmark is shown in Figure \ref{fig:overview}.

Using \textsc{LFQA-E}, we critically assess the efficacy of $17$ evaluation metrics. The experimental results show that current leading evaluation metrics fail to capture core information as human beings from verbose responses when differentiating the better one between two responses with similar quality. Furthermore, we provide analysis on why automatic evaluation metrics fail in LFQA evaluation and find the misalignment between evaluation metrics. Lastly, we try TTRL \citep{zuo2025ttrl} to improve the evaluation performance of model-based metrics and provide some actionable insights.

%% file: main/related_work.tex
\section{Related Work}
\paragraph{Development of LFQA} LFQA \citep{fan2019eli5longformquestion} requires models to generate paragraph-level responses to open-ended questions which is more complex compared to datasets like SQuAD \citep{rajpurkar2016squad100000questionsmachine}, TriviaQA \citep{joshi2017triviaqalargescaledistantly}, and NarrativeQA \citep{kočiský2017narrativeqareadingcomprehensionchallenge}, where answers are primarily words or phrases extracted directly from documents. In LFQA, models must generate comprehensive yet correct responses based on their knowledge or existing evidence documents. Several studies have analyzed the discourse structure of long-form answers \citep{xu2022answercomplexquestionsdiscourse} and have sought to enhance the performance on LFQA \citep{chen2023understandingretrievalaugmentationlongform, akash2023longformquestionansweringiterative}.
\vspace{-2mm}
\paragraph{Evaluation of LFQA} The automatic evaluation of LFQA remains challenging and underexplored. For human annotation, \textsc{Hurdles} \citep{krishna2021hurdlesprogresslongformquestion} and \textsc{WebGPT} \citep{nakano2022webgptbrowserassistedquestionansweringhuman} employ A / B testing, where crowd-sourced annotators are instructed to choose the best of two candidate answers. Since annotation of LFQA requires high expertise, the results of crowd-sourced workers may be unreliable.  To address the gap, \citet{xu2023criticalevaluationevaluationslongform} employs experts for annotation, and tests several evaluation metrics, such as ROUGE \citep{lin-2004-rouge}, BERTScore \citep{zhang2020bertscoreevaluatingtextgeneration}, and BARTScore \citep{yuan2021bartscoreevaluatinggeneratedtext}, on an expert-annotated dataset. Their findings validated that no existing metrics fully align with human judgment. However, the dataset they used lacks expert-written references, which are sourced from Reddit/ELI5, and is limited in scale, comprising only about $260$ samples. More recently, since the development of LLMs and LRMs, many work uses them for evaluation of free-form answers, using prompt \citep{wei2023chainofthoughtpromptingelicitsreasoning}, fine-tuning using specific data \citep{liu2023gevalnlgevaluationusing}, and reinforcement-learning \citep{chen2025rmr1rewardmodelingreasoning}.

%% file: main/methodology.tex
\section{Methodology}
\subsection{Overview}
To reasonably test the evaluation ability of different metrics for LFQA when having a reference, we introduce \textsc{LFQA-E}, a multilingual and comprehensive benchmark composed of different topics and questions. \textsc{LFQA-E} consists of the Chinese version \textsc{LFQA-E-zh} and the English version \textsc{LFQA-E-en}. Table \ref{tab:stats} shows its overview. It includes 1618 questions and 7323 comparisons, consisting of 1193 comparisons in Chinese and 6130 comparisons in English. It spans 15 topics, ranging from history to engineering, ensuring its diversity. \textsc{LFQA-E} comprises expert-annotated references for fair comparison and nuanced responses. Therefore, it is naturally a hard yet reasonable benchmark for LFQA evaluation.

\begin{wraptable}{r}{0.5\textwidth}
\centering
\vspace{-8mm}
\caption{Detailed statistics of \textsc{LFQA-E}. \textbf{Avg Que. Lens}, \textbf{Avg Ref. Lens}, \textbf{Avg Res. Lens} corresponds to question lengths, reference lengths, and response lengths, respectively.}
\scalebox{0.9}{
\begin{tabular}{@{}lcc@{}}
\toprule
  &{\textbf{\textsc{LFQA-E-en}}}&
  {\textbf{\textsc{LFQA-E-zh}}} \\ 
  \midrule 
\textbf{\# Topics} &$ 9$ & $6$ \\
\textbf{\# Questions} & $1026$ & $592$ \\
\textbf{\# Comparisons} & $6130$ &$ 1193$ \\
\textbf{\# Avg Que. Lens} &$ 13.4$ & $24.6$  \\
\textbf{\# Avg Ref. Lens} & $299.1$ & $187.2$ \\
\textbf{\# Avg Res. Lens} &  $245.0$ & $308.3$ \\ 
\textbf{Annotate} & Expert & Expert \\
\bottomrule
\end{tabular}}
\label{tab:stats}
\vspace{-8mm}
\end{wraptable}
\paragraph{Reference-Based Evaluation}
For \textsc{LFQA-E}, references are sourced from academic examinations or widely discussed questions in Reddit/ELI5. After being reviewed by experts with relevant academic backgrounds, these references are ensured to cover all the key points needed to answer the question. This provides a baseline for evaluation metrics to look up and provide a more precise comparison.
\paragraph{Difficult Comparisons}
All the questions contained in \textsc{LFQA-E} have been carefully examined by domain experts to ensure it is answerable and clear to understand. We ensure that models have not seen the data by collecting data from recent examinations and forum questions. The responses are collected from human-written responses, with close scores or upvotes, and model responses generated by comparable LLMs. Therefore, it is hard to distinguish the better one at a glance.

\paragraph{Diverse Benchmark}
We collect $1618$ questions and $7323$ comparisons in $15$ distinct domains, from natural science to social science, to guarantee a diverse and representative benchmark. Also, \textsc{LFQA-E} is multilingual, consisting of examples in both Chinese and English. Moreover, \textsc{LFQA-E} includes three kinds of comparisons, guaranteeing the comprehensibility of the benchmark.

\subsection{Data Processing}
\paragraph{Data Collection}
For \textsc{LFQA-E-zh}, we source our data from CEESQ and PEEQ, where questions, references, and scoring schemas are developed by domain experts, including teachers and professors from high schools and colleges. The records are sampled from 2024 and are based on local examinations restored from PDF files that have not been submitted to online platforms. These questions cover diverse subjects, including politics, history, medicine, psychology, law, and geography. We provide two examples of CEESQ and PEEQ in Appendix \ref{apx:ceesq}. For \textsc{LFQA-E-en}, data is sourced from Reddit/ELI5, where each question is explained without specialized terminology or complex concepts, and we use the top-ranked answer as our candidate reference.
For \textsc{LFQA-E-zh}, to prevent overlap with potential training data, we avoid using data from actual College Entrance Examinations. All problems are sourced from PDF files and have not been uploaded online. Additionally, the questions captured from ELI5 are all from the past six months. We provide a benchmark contamination study in Sec \ref{sub:contamination}. To ensure that all questions are clear and answerable, we instruct GPT-4o with \text{temperature}=0.7 to filter out questions with unclear descriptions. The instruction used is listed in Appendix \ref{sec:inst_data}. We conducted an experiment on the effectiveness of leveraging GPT-4o as a filter, and the results reveal that GPT-4o demonstrates superior performance for this task, achieving 97\% accuracy. Subsequently, to ensure our references contain all information needed to answer the questions, we submit the remaining data for expert annotation. The annotation guidelines are presented in Table \ref{tab:annot_ref}. Each reference was annotated by two annotators, and references were discarded if either annotator labeled them as invalid. The Cohen's kappa coefficient is 0.78, indicating substantial inter-annotator agreement. Through this process, we obtained 1618 questions.

\paragraph{Human Response Collection}
For \textsc{LFQA-E-zh}, we gather examination papers primarily in image format and employ Optical Character Recognition (OCR) systems to extract student responses. Specifically, we choose student answers with close scores to ensure the comparison difficulty. The OCR is conducted using the Volcano Engine API. For \textsc{LFQA-E-en}, we collect responses from the forum section of the corresponding question. Also, we select answers within the many-voted yet close up-votes to make them hard to differentiate. However, the responses we collect for \textsc{LFQA-E-zh} are mainly written during examination, it is concise and well structured, and the responses we collect for \textsc{LFQA-E-en} include some special characters like URLs, which deteriorate our data quality. To handle it, we use GPT-4o to paraphrase and clean our human responses. The instruction we used is shown in Appendix \ref{sec:inst_data}. We randomly sample 100 records and annotate the paraphrased responses to validate the performance of GPT-4o, with the results in Appendix \ref{apx:para}, which indicates that leveraging LLMs doesn't introduce any error. We also provide a case study in Appendix \ref{apx:case_paraphrase}.

\paragraph{Model Response Generation}
When generating model responses, we focus on evaluating whether LLMs can understand the semantic meaning of texts well and properly select the better response. Therefore, we do not impose extremely strict requirements on answer quality. Instead, we ensure the difficulty of \textsc{LFQA-E Bench} by selecting models with similar ranking in the LMSYS Arena \citep{chiang2024chatbot, zheng2023judging, zheng2024lmsyschatm}. On account of responses generated by stronger models like GPT-4o or Claude-4-sonnet will pose a great challenge for our annotators to differentiate, increasing the cost of annotation, we leverage Llama-3-8B-Instruct \citep{dubey2024llama3herdmodels} and GPT-3.5-turbo \citep{openai2023chatgpt} for response generation. For model-generated answers, we use \textit{"Generate reasonable answers to the following questions. Use references or examples if needed"} to prompt LLMs. The generation temperature is set to $1.0$ to encourage diverse and creative responses.
\subsection{Human Annotation}

\paragraph{Annotator Decision}
We hire 10 annotators from relevant aspects or who have taken relevant courses. Then we provide them with clear annotation recipes for better quality control. The annotation recipe is in Appendix \ref{anno}. Each annotator receives $2\$$ for annotating a question, including 4-6 comparisons. To ensure the effectiveness of the annotation, we pre-annotate a subset consisting of 35 records. The annotator will start their work until they reach a 90\% consistency on the subset.

\paragraph{Annotation Setting}
Guided by \citet{xu2023criticalevaluationevaluationslongform}, our evaluation criteria mainly focus on factuality and completeness according to the reference, since almost all responses we collect are already very fluent. Unlike typical A/B testing, our method employs a triple-choice format, giving a tie option, to better capture the subtle differences between answers, as they often show comparable levels of information overlap with the reference while the other information is useless or verbose according to the central topic which can de dropped without hindering the comprehension of the response.
\input{tables/main_results}
\paragraph{Annotation Process}
The annotators assess two responses against a given reference and select the more informative and complete answer or declare a ``tie'' if both are of similar quality. During the process, we treat a piece of information as the basic unit as FActScore \citep{min2023factscorefinegrainedatomicevaluation}. Initially, annotators extract the key information needed to answer the question from the provided reference and check whether the responses under evaluation contain similar statements. Then, they will select a better one based on the overlapped information. To minimize bias and subjectivity, each record is annotated by two independent reviewers. Each comparison takes around $7$ minutes to annotate. For some hard-to-differentiate comparisons, detailed justification is saved to help understand. After annotation, we find the Cohen's kappa correlation of inter-annotator agreement is approximately $0.65$, indicating a substantial agreement. We show a screenshot in Figure \ref{fig:anno}.


%% file: tables/main_results.tex
\begin{table*}[!ht]
\centering
\caption{Performance of evaluation metrics on \textsc{LFQA-E}. The largest value is denoted in \textbf{bold}. We provide the results of other models in Appendix \ref{apx:baseline}.}
\resizebox{.935\linewidth}{!}{
\begin{tabular}{@{}lccccccc@{}}
\toprule
 \multirow{2}{*}{$\mathbf{Model}$} &\multicolumn{2}{c}{\textbf{\textsc{LFQA-E-en}}}&
  \multicolumn{2}{c}{\textbf{\textsc{LFQA-E-zh}}}&
  \multirow{2}{*}{$\mathbf{Avg}_{F1}$} &
  \multirow{2}{*}{$\mathbf{Avg}_{Acc}$} \\   \cmidrule(lr){2-5}&$\mathbf{F1}$ & $\mathbf{Accuracy}$ & $\mathbf{F1}$ &$\mathbf{Accuracy}$      &        \\ 
  \midrule
\multicolumn{8}{c}{Static Evaluation Metric} \\ \midrule
{Human Baseline} & {$77.7$} & {$83.3$} & {$68.9$} & {$76.5$} & {$73.3$} & {$79.9$} \\
Length & {$26.0$} & {$42.8$} & {$33.5$} & {$52.6$} & {$30.8$} & {$47.7$} \\
 ROUGE                &{$37.5$}         & {$55.5$}          & {$34.0$}     & {$49.7$}         & {$35.8$}          & {$52.6$}       \\
 BERTScore    & {$35.9$} & {$54.1$}  & {$36.6$} & {$52.4$} & {$36.3$}   & {$53.3$}  \\
\midrule
\multicolumn{8}{c}{LLMs-based Evaluation Metric} \\ \midrule
 Qwen2.5-32B-Instruct            &{$45.8$} & {$63.5$} & {$41.8$}& {$56.7$}& {$43.8$}  & ${60.1}$ \\
 Qwen2.5-72B-Instruct &$43.1$& {$61.2$} & {$39.0$}& $53.0$& $41.1$  & $57.1$ \\
 Llama3.1-70B-Instruct    &$42.5$& $59.6$ & $29.4$& $30.7$& $36.0$ & $45.2$ \\
 GPT-4o       & $\textbf{46.4}$ & $61.7$ & $42.6$& $53.2$& $\textbf{44.5}$ & $57.5$ \\
 DeepSeek-V3     &{$39.3$}& $57.9$ & $41.1$& {$53.8$}& {$40.2$} & {$55.9$} & \\ 
\midrule
\multicolumn{8}{c}{RM-based Evaluation Metric} \\ \midrule
Skywork-Reward-Llama & $37.3$	&	$54.4$	& $38.2$	&	$53.6$	& $37.8$ & 	$54.0$ \\
Skywork-Reward-Gemma	& $37.5$	& $56.0$ & $33.0$ &$48.3$	& $35.3$ &	$52.2$ \\
RM-R1-Qwen2.5-Instruct-14B	& $36.4$	& $64.9$ & $35.1$ &$51.9$	& $35.8$ &	$58.4$ \\
RM-R1-DeepSeek-Distilled-Qwen-14B	& $43.9$	& $65.7$ & $36.7$ &$53.2$	& $40.3$ &	$59.5$ \\
\midrule
\multicolumn{8}{c}{LRM-based Evaluation Metric} \\ \midrule
o1-mini & $45.9$	& $62.9$	& $\textbf{45.2}$	& $\textbf{58.9}$ & $45.6$	& $\textbf{60.9}$ \\
Deepseek-R1 & $42.9$	& $59.6$	&	$42.4$	&	$57.8$	& $42.7$	& $58.7$ \\
\midrule
\multicolumn{8}{c}{Trained Evaluation Metric} \\ \midrule
Auto-J-6B-bilingual	& $46.0$ &  $\textbf{66.8}$	& $35.4$	&	$51.9$ & $40.7$ & $59.4$\\							
Prometheus-7B-v2.0	& $41.8$	& $64.2$	& $34.1$	& $50.1$ &$ 38.0$ & $57.2$ \\												
M-Prometheus-14B  &  $41.6$ &$ 60.8 $ & $33.9$ & $49.4$ & $37.8$ & $55.1$ \\																			
\bottomrule
\end{tabular}
}
\label{tab:main}
\vspace{-4mm}
\end{table*}

%% file: main/experiments.tex
\section{Experiments}
\subsection{Models}
We evaluate various metrics on \textsc{LFQA-E-en} and \textsc{LFQA-E-zh} respectively, including:
\textbf{Static Metrics:} We use Length-orientation, ROUGE-1 \citep{lin-2004-rouge} and BERTScore (F1) \cite{zhang2020bertscoreevaluatingtextgeneration} since they are widely used as the evaluation metric for LFQA.
\textbf{LLMs:} We select Qwen2.5-32B-Instruct \citep{qwen2025qwen25technicalreport}, Qwen2.5-72B-Instruct, Llama-3.1-70B-Instruct \citep{dubey2024llama3herdmodels}, Deepseek-V3 \citep{deepseekai2025deepseekv3technicalreport}, and GPT-4o \citep{openai2024gpt4ocard}.
\textbf{LRMs:} Considering the high time complexity and cost, we use o1-mini and Deepseek-R1 \citep{deepseekai2025deepseekr1incentivizingreasoningcapability}.
\textbf{RMs:}  We test on Skywork-Reward-Gemma-2-27B-v0.2 \citep{liu2024skywork}, Skywork-Reward-Llama-3.1-8B-v0.2 considering their leading position on Reward Bench \citep{RewardBench}. We also test RM-R1-Qwen2.5-Instruct-14B \citep{liu2024rmr1} and RM-R1-Deepseek-Distilled-Qwen-14B since they represent another paradigm of reward models. We refer to Skywork-Reward-Gemma-2-27B-v0.2 and Skywork-Reward-Llama-3.1-8B-v0.2 as Skywork-Reward-Gemma and Skywork-Reward-Llama for simplification.
\textbf{Evaluation-Specific Models:} There are some models trained to be evaluation models. Among these models, we select Auto-J-Bilingual \citep{li2023generativejudgeevaluatingalignment}, Prometheus-7B-v2.0, and M-Prometheus-14B 
 \citep{kim2024prometheus}. 

\subsection{Implementation Details}
We evaluate all the metrics in both \textsc{LFQA-E-en} and \textsc{LFQA-E-zh}. We use Jieba cut for ROUGE-zh. For BertScore, we use roberta-large for \textsc{LFQA-E-en} evaluation and bert-base-chinese for \textsc{LFQA-E-zh}. We set the temperature at 1.0 for all LLM-based evaluation metrics to encourage diverse responses. We show the results of $\text{temperature} = 0$ in Sec \ref{sec:temp}. The prompts we used are shown in Appendix \ref{sec:inst_eval}. We further conduct a prompt-sensitivity analysis in Appendix \ref{apx:sense}. For models with specific training templates, we adopt them. We include references for models to look up in all our settings. We use accuracy and macro-F1 as our indicators: \begin{equation}
    \text{Acc} = \frac{1}{N}\sum_{i=1}^N \mathbb{I}(\text{pred}_i = \text{label}_i)
\end{equation}
\begin{equation}
    \text{F1}_{\text{macro}} = \frac{1}{||\mathcal{C}||}\sum_{c \in \mathcal{C}} \left(2 \cdot \frac{P_c R_c}{P_c + R_c}\right) 
\end{equation}
where $\mathcal{C} = \{\text{A}, \text{B}, \text{tie}\}$. For LLM-based methods, we include a ``tie'' option in the instruction, while for other methods that return a scalar, we round the scalar to 3 decimal places. For the human baseline, we hire another 3 annotators with doctor’s degrees to ensure the quality. The final scores are obtained by averaging annotators’ results. We use the annotation recipe as previously claimed.
\input{tables/equal_ablate}

\subsection{Main Results}
Table \ref{tab:main} lists our  results. The overall low accuracies and F1-scores of all evaluation metrics indicate the challenge \textsc{LFQA-E} poses to current models and methods. We also provide a cost analysis in Appendix \ref{apx:cost}.
\paragraph{Comparison Between Metrics} Though none of the evaluation metrics achieves a high performance on \textsc{LFQA-E}, we observe that scaling model size doesn't definitely yield a better result. For example, Qwen2.5-32B-Instruct beats Qwen2.5-72B-Instruct by $3\%$.  What's more, LRMs show a great performance compared with LLMs, thanks to their long CoT and extended thinking. RM-based evaluation metrics don't show promising results when generalizing to LFQA evaluation, perhaps because they are trained to give a better one between two responses, renouncing the "tie" option. We will analyze further and give a fairer comparison in Section \ref{sec:fair}.
\paragraph{Comparison Between Indicators} All evaluation metrics struggle to give a tie as good as human beings. Table \ref{tab:equal_ablate} indicates that among the evaluation metrics we test, the best result is just $9.2\%$ for \textsc{LFQA-E-en} and $14.6\%$ for \textsc{LFQA-E-zh}. Observing the responses, we find that they are too conservative to claim two responses are of equal quality. This explains why accuracy is always larger than Macro-F1. The low accuracy on tie comparison reflects the difficulty of \textsc{LFQA-E} again.
\paragraph{Specialized Tuned models Help Boost Performance.} Observing the table above, we find that tuning a base model to be a robust generative reward model helps in LFQA evaluation. Moreover, when changing the base model to a strong reasoning model, the performance gains continually. In addition, SFT models can also achieve performance comparable with models of larger sizes. These phenomenon indicates that tuning is essential for LFQA evaluation.

%% file: tables/equal_ablate.tex

\begin{wraptable}{r}{0.52\columnwidth}
  \centering
  \vspace{-10pt} 
   \caption{Performance of different models on comparisons that humans labeled as \emph{tie}. The largest value in each column is in \textbf{bold}.}
 \footnotesize
  \setlength{\tabcolsep}{4pt} 
  \begin{tabular}{@{}lcc@{}}
    \toprule
    \textbf{Model} & \textbf{\textsc{LFQA-E-en}} & \textbf{\textsc{LFQA-E-zh}} \\
    \midrule
    Deepseek-V3             & 1.8 & 10.2 \\
    Qwen2.5-32B-Instruct    & 7.2 & 7.5 \\
    Qwen2.5-72B-Instruct    & 2.6 & 3.6 \\
    Llama-3.1-70B-Instruct  & 5.0 & 7.7 \\
    o1-mini                 & 7.1 & 14.1 \\
    Deepseek-R1             & 7.4 & 7.2 \\
    GPT-4o                  & \textbf{9.2} & \textbf{14.6} \\
    \bottomrule
  \end{tabular}
  \label{tab:equal_ablate}
  \vspace{-20pt} 
\end{wraptable}

%% file: main/analysis.tex
\section{Analysis}
\subsection{Temperature Matters}
\label{sec:temp}
To ablate the effect of temperature on our validation, we further experiment using a temperature equals $0.0$ for deterministic results. The result in Table \ref{tab:temp_0} shows a noticeable shift in metrics. For instance, the performance of LLM-based Evaluation Metrics such as Qwen2.5-32B-Instruct and GPT-4o are higher compared to when the temperature is set to 1.0. This suggests that deterministic behavior likely reduces the model's variability in responses, leading to more consistent evaluation results. However, the LRM-based evaluation metrics show a significant drop, indicating that models relying on more exploration perform worse or even collapse under deterministic conditions.

\input{tables/api_model_acc_temp_0}
\subsection{Evaluation Metrics Can't Excel at All Settings.}



\begin{figure}[t]
  \centering
  \begin{subfigure}[t]{0.48\columnwidth}
    \centering
    \includegraphics[width=\linewidth]{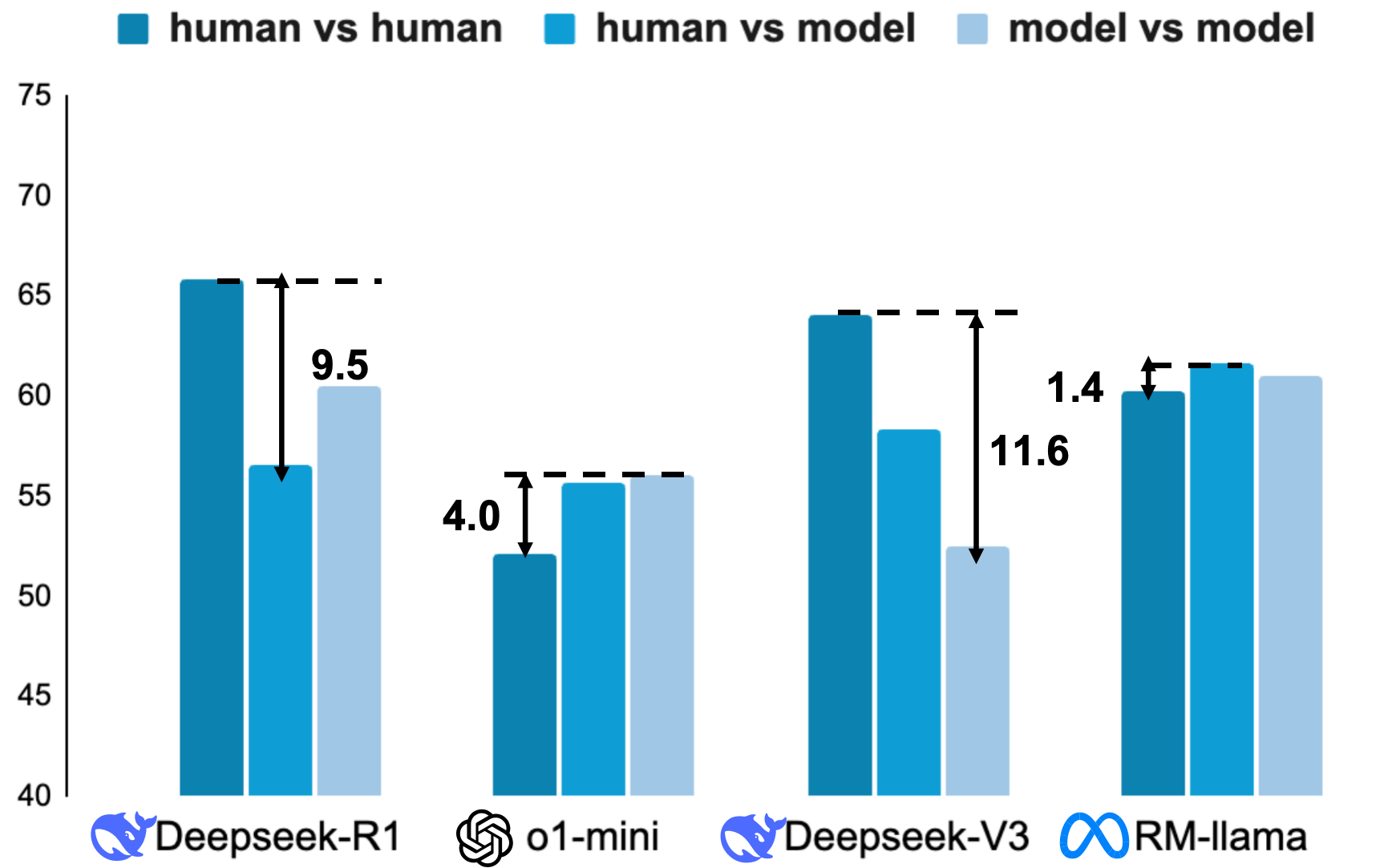}
    \caption{\textsc{LFQA-E-en}}
    \label{fig:lfqa-en}
  \end{subfigure}\hfill
  \begin{subfigure}[t]{0.48\columnwidth}
    \centering
    \includegraphics[width=\linewidth]{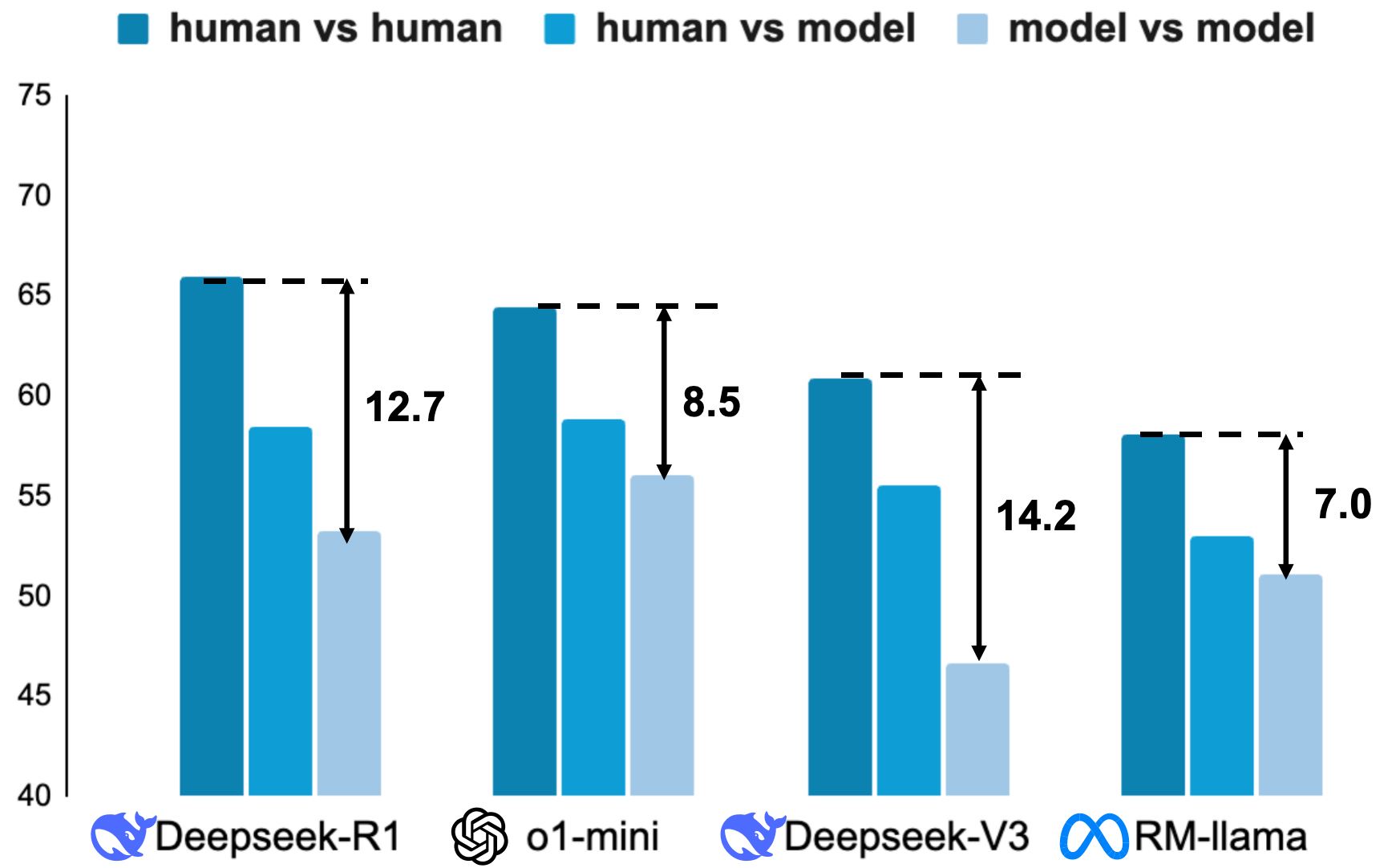}
    \caption{\textsc{LFQA-E-zh}}
    \label{fig:lfqa-zh}
  \end{subfigure}
  \caption{Performance of different models on our three settings on \textsc{LFQA-E}.}
  \label{fig:lfqa}
  \vspace{-3mm}
\end{figure}
To have a thorough understanding of whether the model evaluates human response or model response differently, we experiment on a different group of \textsc{LFQA-E}. We break it into three groups, i.e., \textit{h v. h}, \textit{h v. m}, and \textit{m v. m}, where \textit{h} indicates human response and \textit{m} represents model response, and see the accuracy changes. The results are listed in Figure \ref{fig:lfqa-en} for \textsc{LFQA-E-en} and Figure \ref{fig:lfqa-zh} for \textsc{LFQA-E-zh}. We can observe that for many evaluation metrics, there exists a huge difference between different comparison settings. In \textsc{LFQA-E-en}, the RMs show steady ability while others exhibit degradation when model responses are introduced. In \textsc{LFQA-E-zh}, all the metrics show a drastic accuracy decline under \textit{m v. m}, with a maximum drop of $14.2\%$ from Deepseek-V3. This further validates our assumption that current evaluation metrics can't differentiate between two nuanced responses. we also show the performance across subjects in Appendix \ref{apx:subject}.

\subsection{Reasons Evaluation Metrics fail when Evaluating Long-form Responses.}
\paragraph{For LM-based Evaluation Metric}
We observe the outputs of several LLMs and find that almost all errors arise from the following aspects.
\begin{itemize}
    \item \textit{Keypoints Identification Error}: The model fails to correctly identify and separate bullet keypoints or enumerated lists in responses, leading to poorly structured answers.
    \item \textit{Irrelevant/Incorrect Information Error}: The model does not penalize or filter out irrelevant or factually incorrect details in its responses, reducing accuracy.
    \item \textit{Contradiction Error}: During reasoning, the model generates inconsistent or contradictory statements due to factual hallucinations.
    \item \textit{Formatting Error}: The model produces responses with an improper format.
\end{itemize}
\begin{wrapfigure}{r}{0.45\columnwidth}
    \centering
    \includegraphics[width=\linewidth]{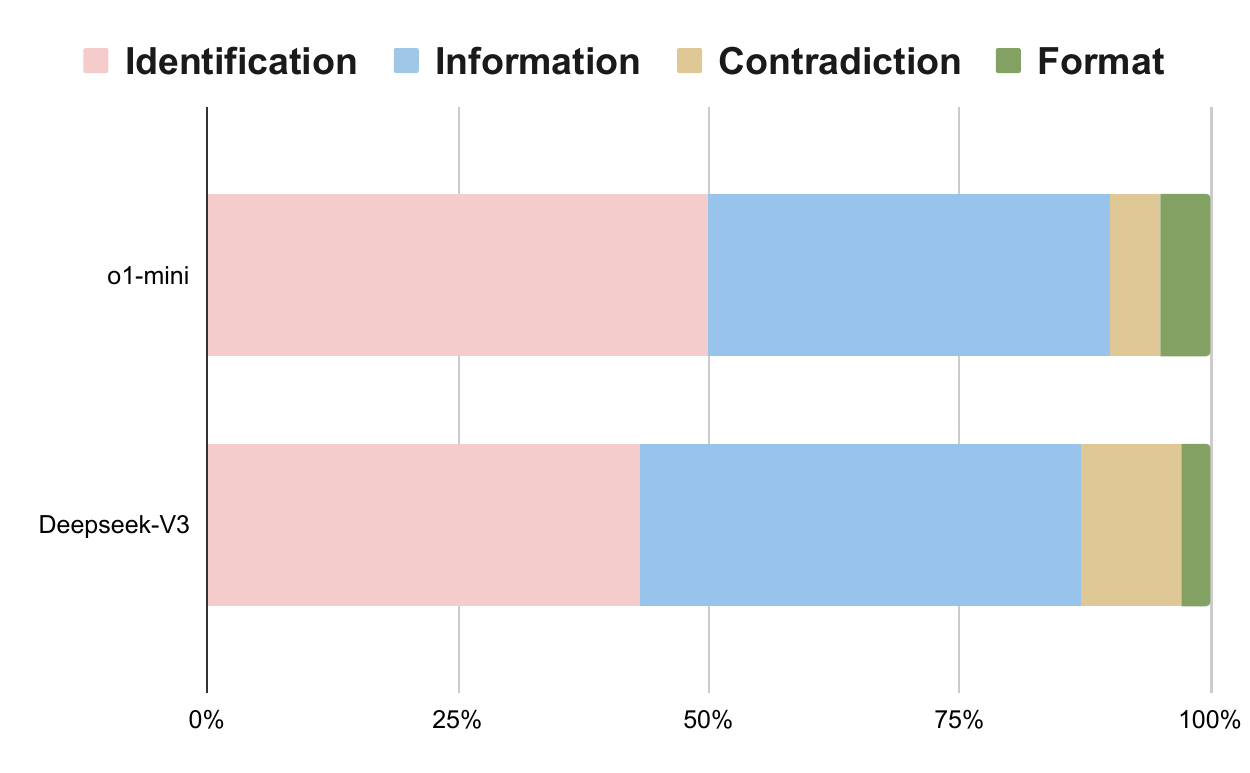}
    \caption{
        Percentage of error types for LMs on the \textsc{LFQA-E} dataset.
    }
    \vspace{-20pt}
    \label{fig:bar}
\end{wrapfigure}
We show the probability of each error occurring in Figure\ref{fig:bar}. We choose Deepseek-V3 and o1-mini for representation. \textit{Point Identification Error} and \textit{Irrelevant/Incorrect Information Error} happen most time, indicating the relatively low inherent ability for LMs when evaluating long-form answers.
\paragraph{Static Evaluation Metrics}
These methods simply leverage word-level or embedding-level similarities, which scratch on surface when evaluating. As described in \citet{fan2024eva}, when considering evaluating two long responses around a topic, there may be many words overlapping. Also, overly long responses dilute semantics, making originally important key information trivial, so metrics fail to consider informativeness, but only focus on similarity.

\begin{figure}[t]
  \centering
  \begin{subfigure}[t]{0.48\columnwidth}
    \centering
    \includegraphics[width=\linewidth]{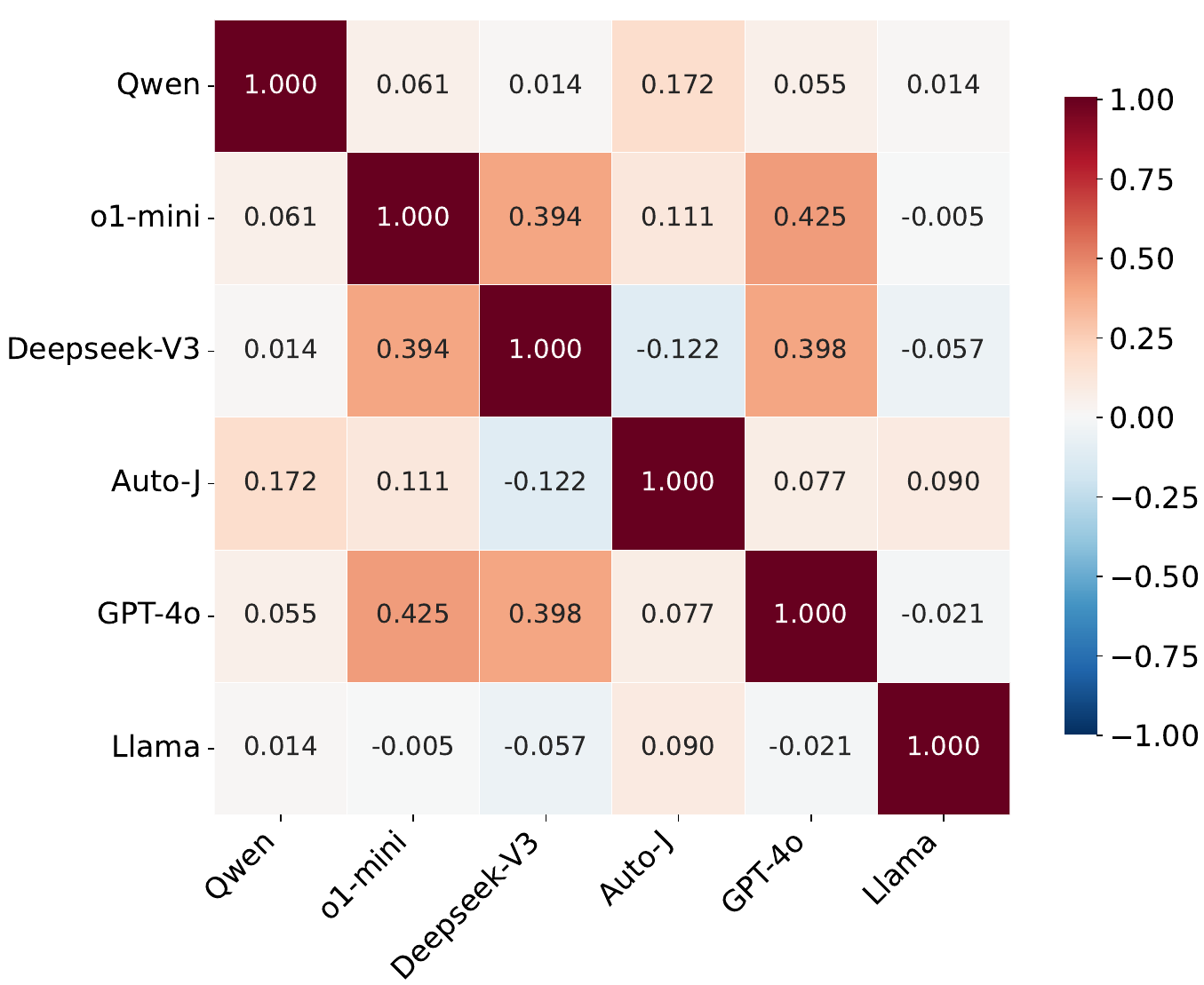}
    \caption{\textsc{LFQA-E-en}}
    \label{fig:corr}
  \end{subfigure}\hfill
  \begin{subfigure}[t]{0.48\columnwidth}
    \centering
    \includegraphics[width=\linewidth]{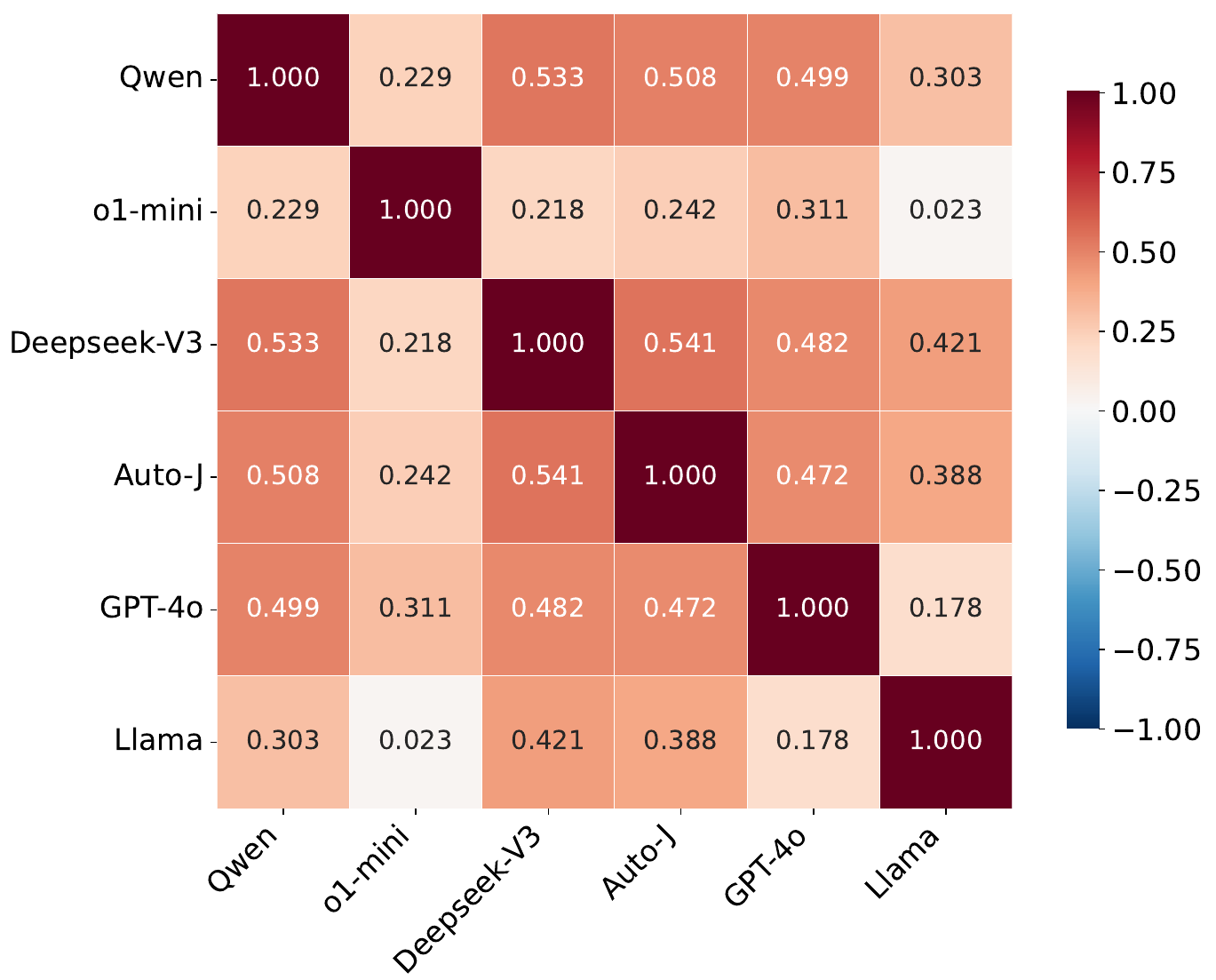}
    \caption{\textsc{LFQA-E-zh}}
    \label{fig:corr_zh}
  \end{subfigure}
  \caption{The Cohen's Kappa Correlation Matrix in \textsc{LFQA-E}.}
  \label{fig:corr_all}
  \vspace{-3mm}
\end{figure}
\subsection{Different Evaluation Metrics don't Agree with Each Other.}
To find whether there is a correlation between different evaluation metrics, we observe detailed evaluation results. Specifically, we select ROUGE, Qwen2.5-32B-Instruct (simplified as Qwen), GPT-4o, Skywork-Reward-Llama (simplified as Llama), o1-mini, and Auto-J-6B-bilingual (simplified as Auto-J), considering their relatively better performance on \textsc{LFQA-E}. Figure \ref{fig:corr} and \ref{fig:corr_zh} show the results. We observe that neither of the two metrics achieves a high correlation, indicating two metrics may contradict each other to a large degree. There are even some negative correlations between the two metrics under \textsc{LFQA-E-en}. This phenomenon further illustrates that there is no stable evaluation result across different metrics.

\subsection{Benchmark Contamination Analysis}
\label{sub:contamination}

To examine whether our evaluation benchmark has potential contamination from pretraining corpora, we conducted perplexity (PPL) and n-gram overlap, the same as \cite{xu2024benchmarking}, using two widely adopted open-source instruction-tuned models: \texttt{Qwen2.5-7B-Instruct} and \texttt{Llama-3.1-8B-Instruct}. The underlying intuition is that if evaluation data is memorized during pretraining, the models would exhibit abnormally low perplexity and high n-gram overlap on the evaluation benchmark.

\paragraph{Perplexity Analysis.}  
Table \ref{tab:contam_ppl} reports the PPL values on both English and Chinese subsets. For \texttt{Qwen2.5-7B-Instruct}, the PPL is 11.60 (en) and 11.73 (zh). For \texttt{Llama-3.1-8B-Instruct}, the PPL is 11.70 (en) and lower at 7.21 (zh), which all fall within tolerable limits and low data contamination potential.


\begin{table*}[h]
\centering
\begin{minipage}{0.45\linewidth}
\centering
\caption{Perplexity (PPL) on the benchmark.}
\begin{tabular}{lcc}
\toprule
Model & en & zh \\
\midrule
Qwen2.5-7B-Instruct & 11.60 & 11.73 \\
Llama-3.1-8B-Instruct & 11.70 & 7.21 \\
\bottomrule
\end{tabular}
\label{tab:contam_ppl}
\end{minipage}
\hspace{0.05\linewidth}
\begin{minipage}{0.45\linewidth}
\centering
\caption{$n$-gram accuracy on the benchmark.}
\begin{tabular}{lcc}
\toprule
Model & en & zh \\
\midrule
Qwen2.5-7B-Instruct & 0.030 & 0.047 \\
Llama-3.1-8B-Instruct & 0.025 & 0.093 \\
\bottomrule
\end{tabular}
\label{tab:contam_ngram}
\end{minipage}
\vspace{-5mm}
\end{table*}

\paragraph{$n$-gram Overlap.}  
We further computed n-gram exact match accuracy between the benchmark and model generations. As shown in Table~\ref{tab:contam_ngram}, the overlap scores remain low across both models, e.g., 0.025--0.030 in English and 0.047--0.093 in Chinese. These values are significantly below contamination thresholds observed in prior work, reinforcing the view that large-scale memorization is unlikely to happen in our benchmark.



\begin{wraptable}{r}{0.5\columnwidth}
\centering
\vspace{-13pt}
\caption{Performance of TTRL.}
\renewcommand{\arraystretch}{1.3}
\begin{tabularx}{\linewidth}{@{} X S[table-format=2.1] @{}}
\toprule
\textbf{Model} & {\textbf{\textsc{LFQA-E-en}}} \\
\midrule
\multicolumn{2}{@{}l}{\textbf{Qwen2.5-3B-Instruct}} \\
\midrule
\quad CoT               & 49.6 \\
\quad Structured Prompt & 59.7 \\
\quad TTRL              & 63.9 \\
\quad TTRL + Clip Higher & 66.5 \\
\midrule
\multicolumn{2}{@{}l}{\textbf{Qwen2.5-7B-Instruct}} \\
\midrule
\quad CoT               & 53.3 \\
\quad Structured Prompt & 60.6 \\
\quad TTRL              & 68.2 \\
\quad TTRL + Clip Higher & 68.6 \\
\bottomrule
\end{tabularx}
\label{tab:ttrl}
\vspace{-12mm}
\end{wraptable}

Overall, both analyses suggest that the benchmark is not heavily contaminated. The English subset appears relatively safe across both models. For the Chinese subset, while Llama-3.1-8B-Instruct shows somewhat lower perplexity and higher n-gram overlap, though n-gram higher than expected, remain within acceptable bounds and don't indicate severe memorization.


\subsection{Comparison with Established Benchmarks}
To demonstrate the value and distinctiveness of our \textsc{LFQA-E} benchmark, we conduct a comparative analysis against several established baselines using frontier models. Our findings consistently show that \textsc{LFQA-E} presents a significantly greater challenge, thereby better differentiating the capabilities of state-of-the-art models.

\paragraph{General LLM Evaluation.}
First, we compare \textsc{LFQA-E} with the expert-annotated benchmark from \citet{xu2023criticalevaluationevaluationslongform}, denoted as ``Expert'' and Feedback-Bench~\cite{kim2024prometheus}. As shown in Table~\ref{tab:comparison_llm}, all leading models achieved substantially lower scores on \textsc{LFQA-E} compared to the baselines. This performance gap validates that \textsc{LFQA-E} probes deeper into the models' reasoning and evaluation capabilities rather than merely surfacing superficial knowledge.

\paragraph{Reward Model Evaluation.}
We further extend our comparison to specialized reward model benchmarks, including RM-Bench~\cite{liu2024rmbenchbenchmarkingrewardmodels} and Reward-Bench~\cite{RewardBench}. The results, presented in Table~\ref{tab:comparison_rm}, reaffirm our conclusion. The consistent performance drop on LFQA-E across various reward models highlights its robustness and difficulty, even for systems specifically designed for evaluation tasks.

\begin{table}[t]
    \centering
    \small
    \caption{Performance comparison between \textsc{LFQA-E} and general evaluation benchmarks.}
    \label{tab:comparison_llm}
    \resizebox{.8\columnwidth}{!}{%
    \begin{tabular}{lccc}
        \toprule
        \textbf{Model} & \textbf{Feedback-Bench} & \textbf{Expert} & \textbf{LFQA-E} \\
        \midrule
        Qwen2.5-32B-Instruct & 86.8\% & 74.2\% & 60.1\% \\
        Qwen2.5-72B-Instruct & 94.4\% & 72.7\% & 57.1\% \\
        GPT-4o               & 89.2\% & 70.0\% & 57.5\% \\
        GPT-5                & 75.6\% & 68.9\% & 62.5\% \\
        DeepSeek-V3          & 83.6\% & 65.8\% & 55.9\% \\
        DeepSeek-R1          & 83.6\% & 68.9\% & 58.7\% \\
        \bottomrule
    \end{tabular}%
    }
\end{table}

\begin{table}[t]
    \centering
    \small
    \caption{Comparison with specialized reward model benchmarks. N/A denotes missing standard results for that benchmark.}
    \label{tab:comparison_rm}
    \resizebox{\columnwidth}{!}{%
    \begin{tabular}{lccc}
        \toprule
        \textbf{Model} & \textbf{Reward-Bench} & \textbf{RM-Bench} & \textbf{LFQA-E} \\
        \midrule
        Skywork-Reward-Gemma-2-27B   & 75.3\% & 69.5\% & 52.2\% \\
        Skywork-Reward-Llama-3.1-8B-v0.2 & 71.8\% & 72.6\% & 54.0\% \\
        Gemini-2.5-flash             & 77.7\% & N/A    & 60.6\% \\
        RM-R1-DeepSeek-Distilled-Qwen-14B & N/A & 71.8\% & 59.5\% \\
        RM-R1-Qwen-Instruct-14B      & N/A & 75.6\% & 58.4\% \\
        \bottomrule
    \end{tabular}%
    }
\end{table}

\subsection{TTRL to Boost Performance}
Here, we try to improve the performance on LFQA-E through prompting and reinforcement learning (RL). Firstly, we use a structured prompt to instruct models to embrace their response within a \texttt{<answer>}...\texttt{</answer>} tag. This leads to a non-trivial performance increase, considering the relatively lower instruction following ability of small language models. Building on this, since RL is widely used to improve the reasoning ability of LLMs \citep{cui2025process, fan2025ssrl}, we use RL to leverage the performance of LLMs on LFQA evaluation. Considering the lack of high-quality data, we implement TTRL \citep{zuo2025ttrl} on our \textsc{LFQA-E-en} as an example. We configure our model with a batch size of 8, a rollout temperature of 1.0, and generate 32 rollouts per prompt. The learning rate is $5e\text{-}7$. During validation, the temperature is 0.0 for consistent results. The reward signal is based on an outcome-based rule, similar to the approach used in Deepseek-R1 \citep{deepseekai2025deepseekr1incentivizingreasoningcapability}. As shown in Table \ref{tab:ttrl}, TTRL yield a substantial performance boost, demonstrating the effectiveness of using RL for this task. Also, the response length grows steadily during the training until the training rewards converge. However, we observe a rapid convergence where all rollouts produce identical preferences, which limits further improvements. We attribute this to the underlying three-category classification which is easily overfitted when the model is over-confident. For a remedy, we implement clip-higher mechanism used in DAPO \citep{yu2025dapoopensourcellmreinforcement}. Also, we observe a performance boost, indicating that when we increase the diversity during rollouts, RL can take effect more stably. However, more sophisticated methods and high-quality data are needed for a better evaluation metric that mimics human preferences.

%% file: tables/api_model_acc_temp_0.tex
\begin{table*}[!ht]
\centering
\caption{Performance of metrics on \textsc{LFQA-E} when temperature is set to 0. The largest value is denoted in \textbf{bold}.}
\resizebox{.935\linewidth}{!}{
\begin{tabular}{@{}lccccccc@{}}
\toprule
 \multirow{2}{*}{$\mathbf{Model}$} &\multicolumn{2}{c}{\textbf{\textsc{LFQA-E-en}}}&
  \multicolumn{2}{c}{\textbf{\textsc{LFQA-E-zh}}}&
  \multirow{2}{*}{$\mathbf{Avg}_{F1}$} &
  \multirow{2}{*}{$\mathbf{Avg}_{Acc}$} \\   \cmidrule(lr){2-5}&$\mathbf{F1}$ & $\mathbf{Accuracy}$ & $\mathbf{F1}$ & $\mathbf{Accuracy}$      &        \\ 
  \midrule
\multicolumn{8}{c}{LLMs-based Evaluation Metric} \\ \midrule
 Qwen2.5-32B-Instruct            &{$46.0$} & {$\textbf{64.5}$} & {$40.0$}& {$54.0$}& {$43.0$}  & ${\textbf{59.3}}$ \\
 Qwen2.5-72B-Instruct &{$41.8$} & {$58.6$} & {$38.5$}& {$52.7$}& {$40.2$}  & ${55.7}$ \\
 Llama3.1-70B-Instruct    &{$36.8$} & {$53.5$} & {$40.5$}& {$50.4$}& {$38.6$}  & ${52.0}$ \\
 GPT-4o       &{$\textbf{49.5}$} & {$63.1$} & {$\textbf{43.9}$}& {$52.7$}& {$\textbf{46.7}$}  & ${57.9}$ \\
 DeepSeek-V3     &{$40.1$}& $57.8$ & $40.9$& {$\textbf{55.1}$}& {$40.5$} & {$56.5$} & \\ 
\midrule
\multicolumn{8}{c}{LRM-based Evaluation Metric} \\ \midrule
o1-mini &{$43.4$} & {$56.2$} & {$4.3$}& {$5.8$}& {$23.9$}  & ${31.0}$ \\
Deepseek-R1 &{2.0} & {2.7} & {32.1}& {45.3}& {17.1}  & ${24.0}$ \\																			
\bottomrule
\end{tabular}
}
\label{tab:temp_0}
\end{table*}

%% file: main/conclusion.tex
\section{Conclusion}
We introduce \textsc{LFQA-E}, a multilingual benchmark for LFQA evaluation. It consists of 1625 questions and 7649 comparisons, spanning 15 topics, from natural science to social science, consisting of 3 settings, i.e., \textit{h v. h}, \textit{h v. m}, and \textit{m v. m}. Each records include a clear question, an authorized reference, and two hard-to-differentiate responses, ensuring its difficulty. We conduct experiments on $15$ automatic evaluation metrics. The results show that none of the metrics can evaluate long-form responses as well as human beings. We further analyze the generalization of different metrics across languages and settings. The results further indicate that all models struggle to generalize well to all comparisons. We find that LRMs and specifically trained evaluation models lead on \textsc{LFQA-E}. The test-time-scaled evaluation model may be used to enhance the performance of LFQA evaluation.

%% file: main/appendix.tex
\section{Use of LLMs}
We use LLMs to refine our writing using Gemini2.5-Pro, GPT-5, and Claude-4.1. We check the refined phrases after generation.

\section{Additional Results}
\subsection{Annotation of LLM Paraphrasing}
\label{apx:para}
We randomly sample 100 records to compare the original responses and the paraphrased responses. We define three types of errors: 1) The paraphrased response contains factual errors; 2) The paraphrased response adds new core points to answer the problem, which may reverse the order; 3) The paraphrased response drops original points or introduces influency or inconsistency. After annotation, we find \textit{error 2} in only 1 out of 100 paraphrased response, indicating the effectiveness of using GPT-4o to paraphrase responses.

\subsection{Prompt Sensitivity Analysis}
\label{apx:sense}
To further investigate the sensitivity to prompt design, we conduct an additional ablation by instructing models to solely provide a final answer without CoT. The results are presented in the table below. This experiment further highlights the importance of CoT for enhancing the ability when evaluating: for most settings under LFQA-E-EN, the performance degrades greatly. While for LFQA-E-ZH, the performance is hard to predict, indicating the language vulnerability.

\begin{table}[h]
    \centering
    \caption{Ablation study on prompt design.}
    \label{tab:ablation_cot}
    \renewcommand{\arraystretch}{1.3} 
    \begin{tabular}{lcccc}
        \toprule
        \multirow{2}{*}{\textbf{Model}} & \multicolumn{2}{c}{\textbf{LFQA-E-EN}} & \multicolumn{2}{c}{\textbf{LFQA-E-ZH}} \\
        \cmidrule(lr){2-3} \cmidrule(lr){4-5} 
         & Acc & F1 & Acc & F1 \\
        \midrule
        
        Qwen2.5-32B-Instruct & $57.3_{\downarrow 6.2}$ & $45.5_{\downarrow 0.3}$ & $52.8_{\downarrow 3.9}$ & $38.6_{\downarrow 3.2}$ \\
        
        Qwen2.5-72B-Instruct & $60.4_{\downarrow 0.8}$ & $45.1_{\uparrow 2.0}$   & $53.9_{\uparrow 0.9}$   & $37.0_{\downarrow 2.0}$ \\
        
        GPT-4o               & $61.4_{\downarrow 0.3}$ & $44.9_{\downarrow 1.5}$ & $51.5_{\downarrow 1.7}$ & $37.7_{\downarrow 4.9}$ \\
        
        DeepSeek-V3          & $51.9_{\downarrow 6.0}$ & $42.2_{\uparrow 2.9}$   & $54.4_{\uparrow 0.6}$   & $37.0_{\downarrow 4.1}$ \\
        
        DeepSeek-R1          & $54.8_{\downarrow 4.8}$ & $41.4_{\downarrow 1.5}$ & $60.0_{\uparrow 2.2}$   & $44.1_{\uparrow 1.7}$ \\
        
        \bottomrule
    \end{tabular}
\end{table}

\subsection{The Ability of Evaluation Metrics Varies across Domains.}
\label{apx:subject}
The performance of evaluation metrics in LFQA-E varies significantly across different domains. As shown in Tables \ref{tab:confounding_factors} and \ref{tab:confounding_factors_en}, models and metrics exhibit distinct strengths and weaknesses depending on the subject area. For instance, on the \textsc{LFQA-E-zh} dataset, models like Qwen2.5-32B-Instruct and GPT-4o consistently excel in subjects such as Geography, Law, and Medicine but perform less effectively in more complex domains like Psychology and History. Conversely, models like DeepSeek-V3 and RM-based Evaluation Metrics show particular strengths in fields like Politics and Law, where the emphasis is on factual accuracy and legal context. Similarly, in the \textsc{LFQA-E-en} dataset, LLM-based models such as Qwen2.5-32B-Instruct perform exceptionally well in Engineering and Technology but show a noticeable decline in subjects like Psychology and Mathematics, possibly due to the more abstract nature of these domains. In contrast, RM-based Evaluation Metrics, such as RM-R1-DeepSeek-Distilled-Qwen-14B, demonstrate impressive performance in fields like Planetary Science and Chemistry, where the data is often more structured and less ambiguous.

\input{tables/confounding_factors}
\input{tables/confounding_factors_en}

\subsection{Results of additional metrics}
\label{apx:baseline}
We incorporate several additional baselines, including factuality-oriented and evaluation-ensemble measures, as shown below in Table \ref{tab:more_baseline}. Through the experimental results, we find that the agentic framework is the most effective method for LFQA evaluation, and majority voting also contributes to improvement.

\begin{table}[h]
    \caption{More baselines on \textsc{LFQA-E}.}
    \label{tab:more_baseline}
    \centering
    \renewcommand{\arraystretch}{1.2} 
    \begin{tabular}{lcccccc}
        \toprule
        \textbf{Model} & \multicolumn{2}{c}{\textbf{LFQA-E-EN}} & \multicolumn{2}{c}{\textbf{LFQA-E-ZH}} & \multicolumn{2}{c}{\textbf{Average}} \\
        \cmidrule(lr){2-3} \cmidrule(lr){4-5} \cmidrule(lr){6-7}
         & Acc & F1 & Acc & F1 & Acc & F1 \\
        \midrule
        
        \multicolumn{7}{c}{\textit{SOTA Models}} \\ 
        \hline 
        Gamini2.5-flash & 63.3 & 45.6 & 57.9 & 44.7 & 60.6 & 45.2 \\
        GPT-5           & 64.8 & 47.7 & 60.1 & 46.7 & 62.5 & 47.2 \\
        \midrule
        
        \multicolumn{7}{c}{\textit{FactScore variants}} \\
        \hline
        FineSurE (GPT-4o) & 47.8 & 35.5 & 42.9 & 32.5 & 45.4 & 34.0 \\
        \midrule
        
        \multicolumn{7}{c}{\textit{Evaluation Ensembles}} \\
        \hline
        ChatEval (GPT-4o) & 70.7 & 48.0 & 49.9 & 40.3 & 60.3 & 44.2 \\
        G-Eval (GPT-4o)   & 68.0 & 50.0 & 55.2 & 42.2 & 61.6 & 46.1 \\
        
        \bottomrule
    \end{tabular}
\end{table}

\subsection{LLMs are Better at Finding Something Better.}
\label{sec:fair}
\input{tables/results_wo_tie} 

Considering that giving a tie option is difficult for both humans and models, we drop out the records that are labeled as a tie and conduct the experiments again. We show the results in Table \ref{tab:results_wo_tie}. After discarding the tied comparison, all the evaluation metrics show nontrivial performance boosts. GPT-4o even gets a $6.4$\% bonus. This increase matches what we find when comparing indicators. Similar to what we observe above, LRMs remain leading on the fairer comparison, and RMs still struggle to generalize to long-form response evaluation. Static evaluation metrics, however, show the least improvement. The experimental results demonstrate the potential of test-time scaling, while reflects the generalization problem of RMs. What's more, specific evaluation models show their great potential once again, ranking first on \textsc{LFQA-E-en}, displaying its future for LFQA evaluation.

\subsection{How to balance the cost and time efficiency between models?}
\label{apx:cost}
To further discuss the tradeoff between efficiency and effectiveness among language models, we calculate the average inference time per question of several LLMs, LRMs and RMs. Table \ref{tab:inference_time} lists the results. Specifically, for the closed-source models that need to invoke api for inference, we calculate their costs on both \textsc{LFQA-E-en} and \textsc{LFQA-E-zh} in Table \ref{tab:api_models_cost}. We observe that \textbf{o1-mini} incurs the highest cost, which is aligned with its relatively higher accuracy and F1 scores. \textbf{DeepSeek-V3}, on the other hand, offers a more cost-effective alternative, though its performance is somewhat lower. For a more practical and cost-efficient evaluation, models like Skywork-Reward-Llama and Skywork-Reward-Gemma have much smaller parameter scales and can be deployed on local resources, showing strong promise for reducing costs while still providing a reasonable evaluation.



\section{Discussion}

We suggest training open-domain RMs or evaluation models, which may help for the evaluation of LFQA, considering their relatively low cost and GPU requirements, but with a decent score. Also, we recommend future work to focus on evaluation workflows that combine the strengths of both LLM-based models and more efficient reward models. Particularly, for open-sourced models, we don't observe steady performance gains as model size scales. Therefore, smaller models with more training data may help more than larger models with some well-designed prompts. 

\section{Case Study}
\subsection{Case from CEESQ and PEEQ}
\label{apx:ceesq}
We provide two cases from CEESQ and PEEQ in Table \ref{tab: ceesq}. We translate them into English for easier comprehension.
\begin{table}[!htbp]
    \centering
        \caption{
  Case Study from CEESQ and PEEQ.
    }
    \scalebox{1.0}{
    \begin{tabular}{@{}p{13.5cm}@{}}
        \toprule
        \textbf{\textsc{Question from PEEQ}:} \\
        What are the classifications and percentages of white blood cells?
        \\
        \midrule
        \textbf{\textsc{Reference}:} \\
        White blood cells are divided into granular cells and non-granular cells. Granular cells include: neutrophils (50\%-70\%); basophils (0-1\%); eosinophils (0.5\%-5\%). Non-granular cells include: monocytes (3\%-8\%); lymphocytes (20\%-40\%) \\
        \midrule
        \textbf{\textsc{Question from CEESQ}:} \\
        Analyze the natural reasons for the numerous sandbars in the Yangtze River estuary area. \\
        \midrule
        \textbf{\textsc{Reference}} \\
        The river has a large discharge volume and carries a large amount of sediment; located at the river estuary, the terrain is low and flat, the flow velocity is slow, with deposition as the main process, leading to massive sediment accumulation; situated at the river-sea interface, tidal backing enhances the deposition process, forming numerous sandbars in the estuary area. \\
        \bottomrule
    \end{tabular}}
    \label{tab: ceesq}
\end{table}

\subsection{Case for Paraphrasing}
\label{apx:case_paraphrase}
We provide a case study from \textsc{LFQA-E-zh} for paraphrasing in Table \ref{tab: case_paraphrase}.
\begin{table}[!htbp]
    \centering
        \caption{
  Case Study for parahrasing.
    }
    \scalebox{1.0}{
    \begin{tabular}{@{}p{13.5cm}@{}}
        \toprule
        \textbf{\textsc{Question from LFQA-E-zh}:} \\
        Reasons for the extremely fragile ecological environment in Guizhou Province.
        \\
        \midrule
        \textbf{\textsc{Original Response}:} \\
        1. Guizhou features karst topography with undulating terrain \\
        2. Serious soil erosion \\
        3. Guizhou Province has few plains and is susceptible to natural disasters \\
        4. Rugged terrain with inconvenient transportation \\
        \midrule
        \textbf{\textsc{Paraphrased Response}:} \\
        The ecological environment of Guizhou Province is extremely fragile for four main reasons: First, Guizhou features karst topography with undulating terrain and complex landforms; Second, serious soil erosion has caused severe depletion of land resources; Third, Guizhou Province has few plains, with most areas being mountainous, making it susceptible to natural disasters; Fourth, the rugged terrain with inconvenient transportation creates certain difficulties for ecological environment protection and restoration. \\
        \bottomrule
    \end{tabular}}
    \label{tab: case_paraphrase}
\end{table}
\input{tables/inference_time_comparison}
\input{tables/api_models_cost}

\subsection{Failed Case}
We show error cases in Table \ref{tab: fail}, Table \ref{tab: fail_2}, and Table \ref{tab:wrong_format_case}. Table \ref{tab: fail} shows the \textit{Incorrect Information Error}. Table \ref{tab: fail_2} shows the \textit{Point Identification Error}. Table \ref{tab:wrong_format_case} shows the \textit{Format Error}.
\subsection{Success Case}
We show successful cases in Table \ref{tab:success} and Table \ref{tab:success_2}.
\input{tables/fail}
\input{tables/fail_2}
\input{tables/wrong_format_case}
\input{tables/success}
\input{tables/success_2}
\section{Instructions}
The following are instructions we used.
\subsection{Data Collection Instructions}
\label{sec:inst_data}
The instruction for the question filter is in Table \ref{tab:llm-filter}. The instruction for the paraphrase is in Table \ref{tab:english-llm-paraphrase}. 

\input{tables/llm_question}
\input{tables/llm_paraphrase}
\subsection{English LLM Evaluation Instruction}
\label{sec:inst_eval}
The instruction for all LLMs and LRMs is in Table \ref{tab:english-llm-eval}. The instruction for Prometheus series is in Table \ref{tab:english-pro-eval}. The instruction for Auto-J is in Table \ref{tab:inst-autoj}.
\input{tables/llm_evaluate_en}
\input{tables/llm_pre} 
\input{tables/llm_autoj}
\section{Annotation}
\label{anno}
\begin{figure*}[h]
    \centering
    \includegraphics[width=0.98\linewidth]{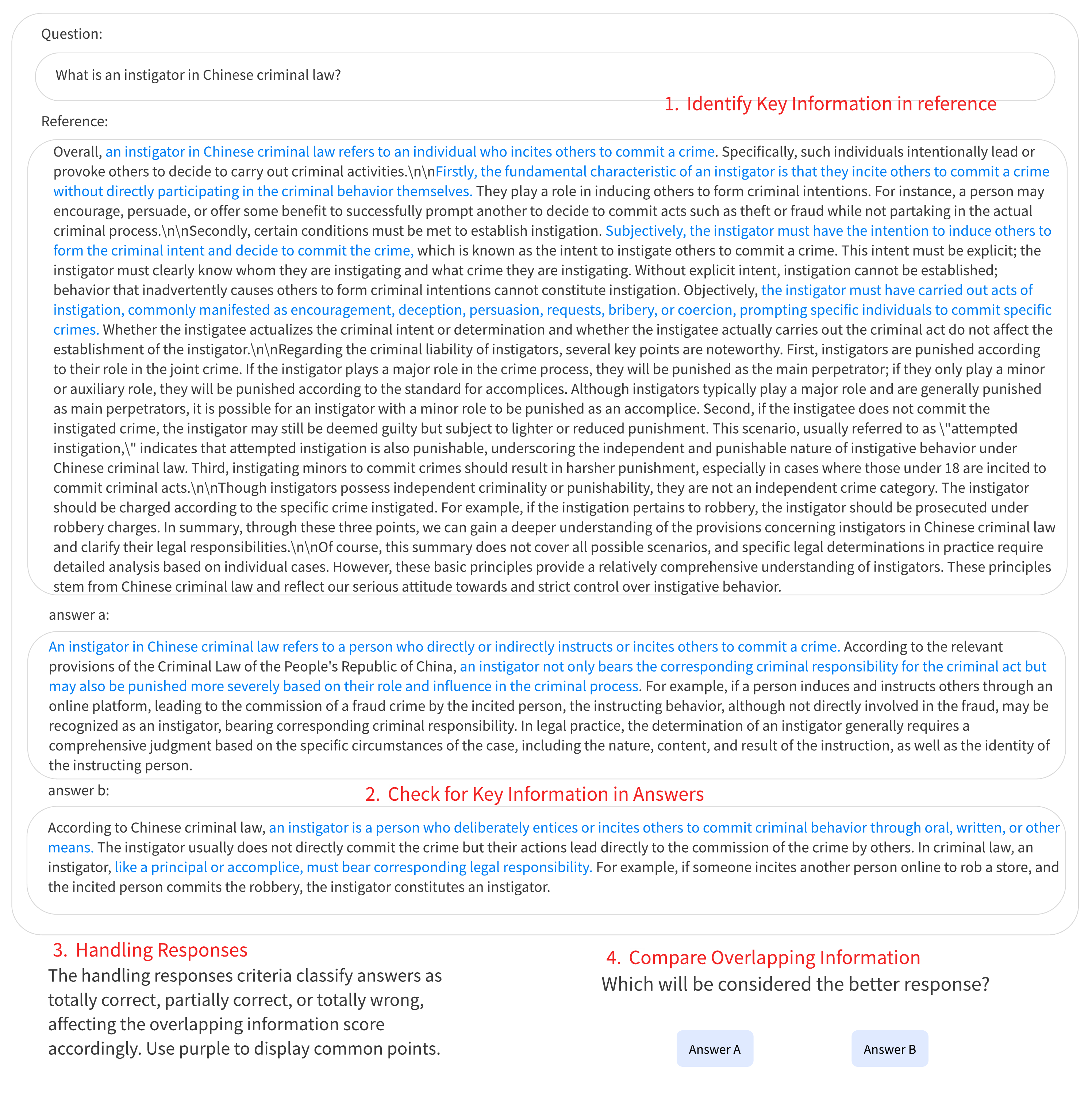}
    \caption{The annotation pipeline.}
    \label{fig:anno}
\end{figure*}

We show the annotation recipe in Table \ref{tab:annot_ref} and Table \ref{tab:annot}. We show a screenshot of annotation pipeline in Figure \ref{fig:anno}.

\input{tables/annot_ref}

\input{tables/annot}

%% file: tables/confounding_factors.tex
\begin{table}[!ht]
\centering
\caption{Results of different topics from \textsc{LFQA-E-zh}. The largest value is denoted using \textbf{bold}.}
\resizebox{\linewidth}{!}{
\begin{tabular}{@{}lcccccc@{}}

\toprule
 \textbf{\textsc{Metrics}} &{\textbf{\textsc{Geography}}}&
  {\textbf{\textsc{History}}} &{\textbf{\textsc{Politics}}}  &{\textbf{\textsc{Psychology}}}& {\textbf{\textsc{Medicine}}}& {\textbf{\textsc{Law}}} \\ 
  \midrule
  \multicolumn{7}{c}{LLM-based Evaluation Metric} \\
  \midrule
  Qwen2.5-32B-Instruct & $64.3$ & $58.2$ & $50.4$ & $44.8$ & $65.1$ &$49.0$\\
  Qwen2.5-72B-Instruct & $\textbf{65.5}$ & $54.3$ & $51.1$ & $42.7$ & $58.3$ &$54.9$\\
  Llama-3.1-70B-Instruct & $41.5$ & $34.4$ & $37.0$ & $46.3$ & $46.6$ &$\textbf{62.6}$\\
  GPT-4o & $62.0$ & $51.8$ & $49.1$ & $44.8$ & $58.3$ &$46.1$\\
  Deepseek-V3 & $60.0$ & $50.5$ & $54.4$ & $40.6$ & $57.3$ &$49.0$\\
  \midrule
  \multicolumn{7}{c}{RM-based Evaluation Metric} \\
  \midrule
  Skywork-Reward-Llama & $58.3$ & $54.1$ & $51.4$ & $34.4$ & $59.2$ &$54.9$\\
  Skywork-Reward-Gemma & $54.3$ & $50.0$ & $41.0$ & $36.5$ & $48.5$ &$56.9$\\
  RM-R1-Qwen2.5-Instruct-14B & $58.6$ & $51.0$ & $45.5$ & $44.8$ & $58.4$ &$54.9$\\
  RM-R1-DeepSeek-Distilled-Qwen-14B & $61.4$ & $51.8$ & $\textbf{62.5}$ & $37.5$ & $54.5$ &$57.8$\\
  \midrule
  \multicolumn{7}{c}{LRM-based Evaluation Metric} \\
  \midrule
  o1-mini & $53.5$ & $46.9$ & $42.5$ & $\textbf{49.0}$ & $\textbf{66.0}$ &$56.9$\\
  Deepseek-R1 & $64.0$ & $\textbf{58.4}$ & $53.2$ & $42.7$ & $67.0$ &$54.9$\\
  \midrule
  \multicolumn{7}{c}{Trained Evaluation Metric} \\
  \midrule
  Auto-J-6B-bilingual & $58.5$ & $55.4$ & $41.7$ & $38.5$ & $58.4$ &$59.8$\\
  Promethus-7B-v2.0 & $54.0$ & $52.6$ & $43.2$ & $40.6$ & $51.5$ &$59.8$\\
  M-Promethus-14B & $56.0$ & $51.3$ & $43.7$ & $41.7$ & $44.7$ &$50.0$\\

\bottomrule
\end{tabular}}
\label{tab:confounding_factors}
\end{table}

%% file: tables/confounding_factors_en.tex
\begin{table}[!ht]
\centering
\caption{Results of different topics from \textsc{LFQA-E-en}. The largest value is denoted using \textbf{bold}.}
\resizebox{\linewidth}{!}{
\begin{tabular}{@{}lccccccccc@{}}
\toprule
 \textbf{\textsc{Metrics}} &{\textbf{\textsc{Engineering}}}&
  {\textbf{\textsc{Biology}}} &{\textbf{\textsc{Technology}}}  &{\textbf{\textsc{Physics}}}& {\textbf{\textsc{Mathematics}}}& {\textbf{\textsc{Economics}}}& \textbf{\textsc{Planetary Science}} & \textbf{\textsc{Chemistry}} & \textbf{\textsc{Other}} \\ 
  \midrule
  \multicolumn{10}{c}{LLM-based Evaluation Metric} \\
  \midrule
  Qwen2.5-32B-Instruct & 59.0 & 64.7 & 59.0 & 52.3 & 55.6 & 63.0 & 64.7 & 64.7 & 65.7 \\
  Qwen2.5-72B-Instruct & 59.8 & 65.7 & 60.2 & 52.2 & 57.1 & 61.9 & 61.6 & 67.5 & 63.4 \\
  Llama-3.1-70B-Instruct & 59.7 & 61.4 & 53.2 & 51.0 & 55.0 & 61.2 & 60.5 & 62.6 & 58.0 \\
  GPT-4o & 60.5 & 65.1 & 60.4 & 55.9 & 58.1 & 63.0 & 64.4 & 64.4 & 65.5 \\
  Deepseek-V3 & 52.6 & 64.9 & 52.8 & 51.9 & 54.4 & 62.7 & 64.4 & 64.7 & 57.1 \\
  \midrule
  \multicolumn{10}{c}{RM-based Evaluation Metric} \\
  \midrule
  Skywork-Reward-Llama & 55.9 & 57.1 & 54.1 & 55.9 & 56.2 & 57.1 & 58.0 & 63.1 & 53.5 \\
  Skywork-Reward-Gemma & 56.9 & 59.0 & 59.2 & 56.3 & 56.8 & 57.0 & 58.5 & 56.6 & 60.0 \\
  RM-R1-Qwen2.5-Instruct-14B & 56.9 & 59.0 & 59.2 & 56.3 & 56.8 & 57.0 & 58.5 & 56.6 & 60.0 \\
RM-R1-DeepSeek-Distilled-Qwen-14B & 65.7 & \textbf{72.1} & 64.5 & 63.9 & 61.3 & \textbf{67.7} & 73.0 & 71.4 & 69.0 \\
  \midrule
  \multicolumn{10}{c}{LRM-based Evaluation Metric} \\
  \midrule
  o1-mini & 56.2 & 64.7 & 53.9 & 54.0 & 58.7 & 64.3 & 67.3 & 62.6 & 57.0 \\
  Deepseek-r1 & 54.0 & 58.6 & 54.3 & 48.3 & 47.1 & 58.8 & 56.3 & 62.6 & 56.1 \\
  \midrule
  \multicolumn{10}{c}{Trained Evaluation Metric} \\
  \midrule
  Auto-J-6B-bilingual & \textbf{72.3} & 68.3 & \textbf{73.1} & \textbf{65.2} & 64.3 & 67.1 & \textbf{67.4} & 69.6 & \textbf{72.9 }\\
  Promethus-7B-v2.0 & 66.8 & 66.4 & 69.0 & 65.1 & \textbf{65.1} & 63.9 & 64.0 & \textbf{72.0} & 57.0 \\
  M-Promethus-14B & 64.1 & 63.5 & 62.0 & 59.5 & 55.2 & 62.1 & 64.9 & 70.6 & 65.5 \\
\bottomrule
\end{tabular}}
\label{tab:confounding_factors_en}
\end{table}

%% file: tables/results_wo_tie.tex
\begin{table}[!ht]
\centering
\caption{Performance of different evaluation metrics on \textsc{LFQA-E}. The examples whose labels are tie are discarded for fairer comparison. The largest value is denoted using \textbf{bold}.}
\scalebox{1.0}{
\begin{tabular}{@{}lccc@{}}
\toprule
 \textbf{\textsc{Model}} &{\textbf{\textsc{LFQA-E-en}}}&
  {\textbf{\textsc{LFQA-E-zh}}} & \textbf{Avg} \\ 
  \midrule
  \multicolumn{4}{c}{Static Evaluation Metric} \\
  \midrule
  Length & $42.7$ & $56.9$ & $49.8$ \textcolor{red}{$\uparrow2.1\%$}  \\
  ROUGE & $57.5$ & $53.8$ & $55.7$ \textcolor{red}{$\uparrow3.1\%$} \\
  BERTScore & $56.0 $& $56.6$ & $56.3$ \textcolor{red}{$\uparrow3.0\%$} \\
  \midrule
  \multicolumn{4}{c}{LLM-based Evaluation Metric} \\
  \midrule
  Qwen2.5-32B-Instruct & $66.8$ & $62.8$ & $64.8$ \textcolor{red}{$\uparrow4.7\%$} \\
  Qwen2.5-72B-Instruct & $63.4$ & $57.6$ & $60.5$ \textcolor{red}{$\uparrow3.4\%$} \\
  Llama-3.1-70B-Instruct & $66.1$ & $34.0$ & $50.1$ \textcolor{red}{$\uparrow4.9\%$} \\
  GPT-4o & $66.3$ & $61.4$ & $63.9$ \textcolor{red}{$\uparrow\textbf{6.4}\%$} \\
  Deepseek-V3 & $60.0$ & $60.0$ & $60.0$ \textcolor{red}{$\uparrow4.1\%$} \\
  \midrule
  \multicolumn{4}{c}{RM-based Evaluation Metric} \\
  \midrule
  Skywork-Reward-Llama & $56.7$  &  $58.4$ & $57.6$ \textcolor{red}{$\uparrow3.6\%$}\\
  Skywork-Reward-Gemma &  $58.2$ & $52.5$ & $55.4$ \textcolor{red}{$\uparrow3.2\%$}\\
  RM-R1-Qwen2.5-Instruct-14B	& $67.5$	& $56.2$ & $61.9$ \textcolor{red}{$\uparrow3.5\%$}  \\
RM-R1-DeepSeek-Distilled-Qwen-14B	& $68.0$	& $57.5$ &$62.8$ \textcolor{red}{$\uparrow3.3\%$} \\
  \midrule
  \multicolumn{4}{c}{LRM-based Evaluation Metric} \\
  \midrule
  o1-mini & $67.3$ & $\textbf{64.2}$ & $\textbf{65.8}$ \textcolor{red}{$\uparrow4.9\%$} \\
  Deepseek-R1 & $61.6$& $63.1$ & $62.4$ \textcolor{red}{$\uparrow3.7\%$} \\
  \midrule
  \multicolumn{4}{c}{Trained Evaluation Metric}\\
  \midrule
  Auto-J-6B-bilingual & \textbf{70.0} & 57.3 & 63.7 \textcolor{red}{$\uparrow4.3\%$} \\
  Prometheus-7B-v2.0 & $66.5$ & $54.1$& $60.3$ \textcolor{red}{$\uparrow3.1\%$} \\
  M-Prometheus-14B & $63.7$ & $53.3$&  $58.5$ \textcolor{red}{$\uparrow3.4\%$}\\

\bottomrule
\end{tabular}}
\label{tab:results_wo_tie}
\end{table}

%% file: tables/inference_time_comparison.tex

\begin{table}[!ht]
\centering
\caption{The average inference time of different language models per question, arranged in descending order.}
\scalebox{0.95}{
\begin{tabularx}{0.7\textwidth}{@{}l>{\centering\arraybackslash}X@{}}
\toprule
\textbf{Model} & \textbf{\textsc{Time (seconds)}} \\ 
\midrule 
Skywork-Reward-Llama     & 0.21 \\ 
Qwen2.5-32B-Instruct     & 0.27 \\
Skywork-Reward-Gemma     & 0.64 \\
Qwen2.5-72B-Instruct     & 1.03 \\
o1-mini                  & 1.80 \\
GPT-4o                   & 3.56 \\
DeepSeek-V3              & 12.00 \\
DeepSeek-R1              & 14.13 \\
\bottomrule
\end{tabularx}}
\label{tab:inference_time}
\end{table}

%% file: tables/api_models_cost.tex


\begin{table}[!ht]
\centering
\caption{Overall cost on closed-source models using API. The largest value is denoted in \textbf{bold}.}
\scalebox{0.90}{
\begin{tabularx}{0.8\textwidth}{@{}l>{\centering\arraybackslash}X>{\centering\arraybackslash}X@{}}
\toprule
\textbf{Model} & \textbf{\textsc{LFQA-E-en}} & \textbf{\textsc{LFQA-E-zh}} \\ 
\midrule 
DeepSeek-V3   & 3.6\$  & 0.7\$  \\ 
DeepSeek-R1   & 18.0\$ & 6.9\$  \\
o1-mini       & \textbf{67.0}\$ & \textbf{14.2}\$ \\
GPT-4o        & 35.6\$ & 5.7\$  \\
\bottomrule
\end{tabularx}}
\label{tab:api_models_cost}
\end{table}

%% file: tables/fail.tex
\begin{table*}[!htbp]
    \centering
        \caption{
  Case Study from \textsc{LFQA-E} from GPT-4o. This results from an incorrect information error.
    }
    \scalebox{1.0}{
    \begin{tabular}{@{}p{13.5cm}@{}}
        \toprule
        \textbf{\textsc{Question}:} \\
        How does law enforcement track down a criminal during a manhunt? 
        \\
        \midrule
        \textbf{\textsc{Reference}:} \\
        1) Do we know where he/she is likely to “go to ground?” We try to find people he/she relies on from phone taps, social media, prior crimes etc. You’d be amazed how many guys are found at \textcolor{blue}{their aunt’s/ex’s/former cellmate’s place}. \\
        2) Put the word out to \textcolor{blue}{local TV news}. In my experience they plaster the manhunt news and mugshot on their website landing page right away. \\
        3) License plate scanner hits on \textcolor{blue}{patrol cars and traffic cameras}, if available. This is surprisingly effective. \\
        4) \textcolor{blue}{Social media phishing}. Is he/she into certain drugs? Certain types of prostitutes? Is he trying to find his wife or kids who are working with victim services and in hiding (very common in bad domestic abuse cases)?  \\
        \midrule
        \textbf{\textsc{Response 1}:} \\
        There are a lot of ways that law enforcement can track someone during a manhunt. First and foremost, we live in the age of technology, and technology is a really powerful tool for tracking someone. A cell phone is pretty much always transmitting your approximate location, and law enforcement can gain access to the satellite that keeps track of that location. \\
        There are also a few tried and true methods: the public. Law enforcement will almost always provide a physical description of a suspect, complete with a photograph or artist rendition, and urge people to report anyone who fits the bill close enough to be the person. This is a powerful tool when to get anything modern, you need to interact with other humans or come into close contact with human society. \\
        There are also recordings, photographs, and footage captured by \textcolor{blue}{CCTV cameras everywhere}. Police may also deploy tracking hounds if they know who you are and have anything that has your smell on it and escaping on foot from persistent, trained hunting dogs is unbelievably difficult." \\
        \midrule
        \textbf{\textsc{Response 2}:} \\
        Ohhhhh so many ways. Really depends on what resources are on hand.\\
        1. \textcolor{blue}{Suspect probably has friends and family.} Check their houses, interview them, see what they know about the suspects hobbies and motives. Do they have a goal they're trying to achieve? Is that goal local or would it send them running elsewhere?\\
        2. Flowing from above, roadblocks. If you generally know where your suspect is, where they might try to go, and what means of transportation are available to them, you can monitor those points of access until they show up.\\
        3. Technology. Where did the suspect use their credit card. Oh, they pulled out cash to avoid being traced? We can see where they made the withdrawal and go from there. Is their phone still online? Can we call and trace them? \textcolor{blue}{Were they recently active on social media? How about their friends?} A lot of those posts have geotags in the metadata \\
        \midrule
        \textbf{\textsc{Model Judge}:} \\
        - Response 1 aligns well with the standard answer by emphasizing technology (cell phone tracking), public involvement (providing descriptions and photos), and surveillance methods (CCTV cameras and tracking dogs). It touches upon finding the suspect's connections, similar to the standard answer's first point about identifying likely places to "go to ground." \\
        - Response 2 also covers relevant points such as investigating friends and family, setting up roadblocks, and using technology like phone tracing and social media analysis. However, it delves into areas like credit card tracking, which the standard answer specifically mentions as ineffective. \\
        Overall, \textbf{\textcolor{red}{Response 1}} more closely mirrors the strategies outlined in the standard answer. (\textit{\textcolor{cyan}{The Judgement is \textbf{incorrect} by incorrectly estimating the weights of irrelevant or unimportant keypoints.}}) \\
        \bottomrule
    \end{tabular}}
    \label{tab: fail}
\end{table*}

%% file: tables/fail_2.tex
\begin{table*}[!htbp]
    \centering
        \caption{
  Case Study from \textsc{LFQA-E} from GPT-4o. This results from an identification error.
    }
    \scalebox{1.0}{
    \begin{tabular}{@{}p{13.5cm}@{}}
        \toprule
        \textbf{\textsc{Question}:} \\
        Why do swear words exist? What is it about swear words that make us feel better? Why don't other normal words make us feel the same way swear words do? \\
        \midrule
        \textbf{\textsc{Reference}:} \\
        Swear words fall broadly into two categories: \textcolor{blue}{profanity and obscenity}.Profanity originates with the Ten Commandments, specifically "Do not take the Lord's name in vain" in Abrahamic religions. Generally the idea though is "don't disparage or devalue the sacred". This is where the term "curse words" comes from, as you would curse someone by saying "God damn you". It's also where the term "swear words" comes from, i.e. "I swear to God". Obscenity, by contrast, is primarily words that \textcolor{blue}{relate to sex} or \textcolor{blue}{using the toilet}. These are activities during which you would be vulnerable to being attacked, and also that produce excrement that could be harmful for others. There is sometimes a need to discuss these things in polite company, \textcolor{blue}{so terms were divided into "polite" and "rude"}. \textcolor{blue}{In English}, this distinction largely comes from the language of origin. English has linguistic roots in \textcolor{blue}{both Romance languages (ones based on Latin like French) and Germanic languages}. At the time when the language was developing, \textcolor{blue}{higher society had more background in Romance languages while the lower classes were more Germanic}. So the former words were considered refined and the latter crude and crass. Most English obscenities have close analogs in German. In modern usage, there's actually official definitions of profanity and obscenity by the FCC. Profanity is using these words in an \textcolor{blue}{exclamatory sense}, while obscenity is using these words to \textcolor{blue}{describe the actions}. The former is more allowable than the latter. As for why it feels good to swear, \textcolor{blue}{it's mostly because breaking a taboo gives a feeling of exhilaration}, and if nobody actually suffers as a result, you also get to feel like you're "getting away with it". Basically, if we didn't put so much emphasis on NOT saying the words, it wouldn't feel so good when we do say them.\\
        \midrule
        \textbf{\textsc{Response 1}:} \\
        Language is a tool of communication. Expletives communicate something that you basically cannot express using polite language. There's certainly an element to them where the social transgression is part of what makes them function as they do. What is or isn't considered a swear word largely depends on cultural values. \textcolor{blue}{In the Middle Ages it's pretty apparent that words related to bodily functions or sexuality were not considered particularly offensive}, while blasphemies were generally considered a lot more offensive than they are in most modern societies. Medieval folks appear to have been simultaneously less prudish but more fearful of the almighty. \\
        \midrule
        \textbf{\textsc{Response 2}:} \\
        Language has been used to separate classes for a very long time. Do you know why a pig is a pig in the pen but pork on your plate? Or why a cow becomes beef when you serve it up?  Because one word was used by the lower class who raised the animals and the other word was used by the upper class who could afford to consume it (in this case, \textcolor{blue}{Normans}). \textcolor{blue}{Words used by the lower classes frequently, but not by upper classes}, will be found to be taboo in the upper classes, leading them to be essentially banned in polite society.  They have power because they are taboo, and for no other reason.  The word 'merde' has no power to you unless you speak French, and the word 'crap' has no power to you unless you're 6. \textcolor{blue}{You've been conditioned to believe these words are strong, so you feel strong when saying them}. It really is that simple. It's nothing more than the placebo effect.\\
        \midrule
        \textbf{\textsc{Model Judge}:} \\ 
        Upon evaluating both responses against the reference answer:\\
        - Response 1 aligns closely with the standard answer by discussing the role of swear words in communication, the significance of social transgression, and cultural variations in what is considered offensive. It also touches on historical perspectives, similar to how the reference answer differentiates profanity and obscenity based on origins and societal values.\\
        - Response 2 introduces the concept of language separating classes and the power of taboo stemming from class distinctions. While it addresses the conditioning and perceived strength of swear words, it deviates more from the primary focus of the standard answer, which centers on cultural and linguistic origins rather than class-based language differentiation.\\
        Therefore, \textbf{\textcolor{red}{Response 1}} is better.(\textit{\textcolor{cyan}{The Judgement is \textbf{incorrect} by observing the number of information coexisting.}}) \\
        \bottomrule
    \end{tabular}}
    \label{tab: fail_2}
\end{table*}

%% file: tables/wrong_format_case.tex
\begin{table*}[!htbp]
    \centering
        \caption{
  Case Study For \textsc{LFQA-E} from GPT-4o. This results from a format error.
    }
    \scalebox{1.0}{
    \begin{tabular}{@{}p{13.5cm}@{}}
        \toprule
        \textbf{\textsc{Question}:} \\
        Can someone explain Benford's Law. I get that certain numbers show up more often in large data sets, but why? \\  
        \midrule
        \textbf{\textsc{Reference}:} \\
        Benford's Law is typically stated about data that span multiple orders of magnitude; so we have values that are many times larger than others instead of them being close together. Next is that we have at least a bit of what is called scale-invariance: unlike many simpler probability problems, we don't assume that all numbers are equally likely, but instead that larger ones are rarer. More precisely, we want roughly the same chance to land between x and 2x as for landing between y and 2y. If those two are satisfied, Benford's Law states that the leading digit 1 is more likely than an other, and the larger digits are the less common ones. There can be no perfect examples of scale-invariance in probability. But between those lie dozens of orders of magnitude, easily more than enough within reason...\\
        \midrule
        \textbf{\textsc{Response 1}:} \\
        Benford's law says if you gather up a bunch of random numbers that appear "in the wild" and you tally up the first digits of each of these numbers then you will tend to find lots of 1's and not many 9's. This tends to happen when you have a range of numbers which is spread out over multiple orders of magnitude. The reason is when numbers are spread out over many orders of magnitude, it tends to be roughly uniform when measured on a "log scale". This means that the number of data points between x and 2x will be about the same for different values of x. In other words, there should be as many values between 100 and 200 as there are between 200 and 400. \\
        \midrule
        \textbf{\textsc{Response 2}:} \\
            Benford's law says the first digit of a number picked from a large range of numbers tends to be a 1. Think of it like this between 1 and 20, 11 of the 20 numbers start with a 1. Between 1 and 99, 11 of the 99 numbers start with a 1. This repeats for 0-200 and 0-999 etc always holding the max of "just over half" and the min of "about 11 percent" so if you average that for all ranges, you get that about 30 percent of numbers in an unknown range start with 1. Between 1 and 89, 1 number starts with 9, basically 1/89 percent. Going up to 99 brings us back to 11/99, but now 11 is the max and "almost 0" is the min, so again average it over all ranges, and you get more like 4 percent of lead numbers. You can then apply this to some fraud cases. If the numbers span multiple orders of magnitude AND should be roughly random, AND there are a lot of them, you should expect them to match Benford's law pretty well. If they don't, one of the 3 requirements is probably false. If you know the first and last are true, you can say "these probably arent actually random".\\
        \midrule
        \textbf{\textsc{Model Judge}:} \\
            - Strengths of Response 1: \\
              Correctly identifies that Benford's Law applies to numbers spanning large ranges. \\
              Explains the higher probability of 1 as a leading digit by comparing intervals which aligns with the scale-invariance idea. \\
              Mentions the decreasing probability for higher digits. \\
              Correctly notes the conditions for applying Benford's Law and its use in fraud detection.\\
            - Weaknesses of Response 1: \\
              Doesn't explicitly mention scale-invariance, though the intuition is somewhat captured. \\
              The explanation for why 1 is more likely is a bit simplistic.\\
            - Strengths of Response 2:
              Introduces Benford's Law in a simple, relatable context.
              Correctly notes that 1 and 2 are more common as leading digits.\\
            - Weaknesses of Response 2: \\
              Misrepresents the core reasons for Benford's Law. The explanations given are incorrect or misleading. Benford's Law is not about human behavior but about inherent properties of scale-invariant data. \\
              Doesn't mention orders of magnitude or scale-invariance at all, which are central to the reference answer. \\
              The references at the end are irrelevant to the actual content of the response.\\
              \textit{\textcolor{blue}{No preference provided!}} \\
        \bottomrule
    \end{tabular}}
    \label{tab:wrong_format_case}
\end{table*}

%% file: tables/success.tex
\begin{table*}[!htbp]
    \centering
        \caption{
  Case Study For \textsc{LFQA-E} from Deepseek-V3.
    }
    \scalebox{1.0}{
    \begin{tabular}{@{}p{13.5cm}@{}}
        \toprule
        \textbf{\textsc{Question}:} \\
        how do metals put together with electricity cause technology and chips and computers and data etc? \\  
        \midrule
        \textbf{\textsc{Reference}:} \\
        \textcolor{blue}{Silicon along with a little bit of some other elements} combined form what is called a semiconductor, of which there are 2 types, the n type and the p type. When you then \textcolor{blue}{sandwich these semiconductors in a specific way, it has the property that applying a current to these sandwiches causes it to become conductive}, or if you invert the sandwich it has the property that it's conductive when you don't apply a current and then not conductive when you do. So essentially, what you have created is an electrical switch, which is controlled by electricity, and we call these transistors.\\
        If you're looking at how these transistors then form a computer, you can \textcolor{blue}{arrange transistors to perform logical functions}. A simple one would be the \textcolor{blue}{NOT gate}, which just uses the second switch I mentioned that turns off when there is an input. An \textcolor{blue}{AND gate} would be if you took two switches, placed them one after the other, and each switch is then controlled by an input. Both inputs need to be on for there to be an output. \\
        \midrule
        \textbf{\textsc{Response 1}:} \\
        For that, you need to look at boolian values and at turing machines. A boolian value is just a yes no answer. Yes vs no, on vs off, one vs zero. Technically you use boolian values all the time in speech. “Is Tommy in the bathroom?” You could answer “Yes”, but you could also say “one”, and anyone who knows Boolian will understand. \\
        A Turing machine is just a name for any machine that uses Boolian values. It has rules, for example “if you see seven zeros in a row followed by a one, then do”. It seems pretty abstract. But math people have actually proved that you can do any calculation, think any thought, perform any task, if you have the right rules designed for your turing machine. \\
        So once we figured that out, it just became a matter of inventing the right rules for the machine, and inventing the best form for a machine to store boolian values. \textcolor{blue}{A transistor is always either a One or a Zero}, and they can be made very small, so it makes sense to use them for your Turing Machine. \\
        \midrule
        \textbf{\textsc{Response 2}:} \\
            When metals are put together with electricity in the context of technology, it typically refers to the process of creating integrated circuits or microchips. These chips are made up of layers of different metals, such as silicon, aluminum, and copper, that are carefully arranged on a silicon wafer. \\
            Electricity is used to power the various components on the chip, allowing them to perform their specific functions. \textcolor{blue}{The flow of electricity through these metal layers enables the chip to process and store information}, ultimately allowing computers and other electronic devices to function.\\
            Additionally, metals such as gold and copper are used in the construction of circuit boards and connectors within electronic devices. These metals have excellent conductivity properties, which allow for the efficient flow of electricity throughout the device.\\
        \midrule
        \textbf{\textsc{Model Judge}:} \\
            Response 1 delves into the concept of Boolean values and Turing machines, which extends the topic to how computers "think" and process data. It links the use of transistors to Turing machines and emphasizes the binary nature of data processing, which aligns well with the reference answer's explanation of logical functions and binary systems accessible through transistors.\\
            
            Response 2 focuses on the physical aspects of creating chips, such as the use of various metals and their conductivity. While it provides a general overview of chip construction and functionality, it lacks depth on the conceptual workings, like the function of transistors and binary systems, which is a significant aspect of the reference answer.\\
            
            The reference answer provides an explanation centered around the role of transistors as switches and their organization to perform logical operations. It links this to the representation of numbers and data processing in computers through binary systems, which Response 1 addresses more comprehensively compared to Response 2.\\
            
            Therefore, \textcolor{magenta}{\textbf{Response 1}} is better. \textit{\textcolor[RGB]{202,12,22}{(The evaluation is correct!)}} \\
        \bottomrule
    \end{tabular}}
    \label{tab:success}
\end{table*}

%% file: tables/success_2.tex
\begin{table*}[!htbp]
    \centering
        \caption{
  Case Study For \textsc{LFQA-E} from Deepseek-V3.
    }
    \scalebox{1.0}{
    \begin{tabular}{@{}p{13.5cm}@{}}
        \toprule
        \textbf{\textsc{Question}:} \\
        How did human calculate time?  \\  
        \midrule
        \textbf{\textsc{Reference}:} \\
        For a long time we didn't and no one cared. However, once you start farming, you need to know what season it is, what hour of the day it is more or less, so you know what you can and should do on the fields for a bountiful yield. The easiest way to do is to \textcolor{blue}{watch the celestial bodies, namely the two major ones, the Sun and the Moon} (the Stars also helped, but eventually it was more of a tool for navigation). Over time people figured out that the Sun shifts, days are not evenly distributed over the year, but luckily it goes in cycles, so it is predictable. Once you make a chart of these shifts throughout the year, you got yourself more or less a \textcolor{blue}{calendar}. \textcolor{blue}{Civilizations with accurate-enough calendars survived and thrived}. Calendars got more and more accurate over time as our math and science got better and here we are, keeping time down to \textcolor{blue}{split seconds}. By the way, it does not fit perfectly. We accumulate quite the error over time, but the Gregorian Calendar is designed as a way to correct for these accumulative errors. The Persian Solar Calendar is the most accurate one, because the new year starts after a celestial event. Gregorian calendar has a bigger focus on religious events rather than any celestial event, so it needs corrections. We have a very accurate time measurement, but that is NOT what is used on the daily, because we don't need it.\\
        \midrule
        \textbf{\textsc{Response 1}:} \\
        Absolutely nothing about the way humans decided to calculate time “fits neatly” into Earth’s behaviours. Years are actually 365 days plus 6 hours and 9 minutes, days are actually 23 hours and 56 minutes long.Ot her time systems have been attempted, with the Romans dividing day and night into 12 hours, which \textcolor{blue}{varied in actual length as the days and nights shifted} through the year. They also had the “Civil Day” system: formal names given to time according to certain markers (such as separate named times for “rooster is crowing” and “rooster stops crowing”, or “time we light candles” and “bed time”). Humans have been struggling with measuring it forever. \\
        \midrule
        \textbf{\textsc{Response 2}:} \\
            Time has been calculated and measured in various ways throughout history. One of the earliest methods used by ancient civilizations was \textcolor{blue}{observing the movement of celestial bodies such as the sun, moon, and stars}. By tracking the positions of these objects in the sky, early astronomers were able to develop \textcolor{blue}{calendars and timekeeping systems}. The modern concept of time measurement is based on the International System of Units (SI). In terms of dividing the day into \textcolor{blue}{hours, minutes, and seconds}, this system was developed by the ancient Babylonians and Egyptians. The Babylonians used a base-60 system for counting, which led to the division of the hour into 60 minutes and the minute into 60 seconds. The Egyptians also used a similar system of timekeeping. Overall, \textcolor{blue}{the calculation of time has evolved over centuries through a combination of astronomical observations, scientific advancements, and cultural practices}. The precise measurement of time today is achieved through the use of atomic clocks and coordinated universal time (UTC) to ensure accurate timekeeping worldwide.\\
        \midrule
        \textbf{\textsc{Model Judge}:} \\
            The reference answer emphasizes the historical development of time measurement, focusing on the transition from no precise timekeeping to the necessity of using celestial bodies for farming. \\

            Response 1 discusses the inconsistency and challenges of time measurements historically, noting the Roman system and difficulties faced by humans in tracking time over time. While it touches on early attempts to measure time, it doesn't delve specifically into celestial observations or the evolution of calendars as much.\\

            Response 2 covers the historical aspect of using celestial bodies and tracks the progression to modern timekeeping with atomic clocks. It also discusses the Babylonian and Egyptian contributions to dividing time. This response aligns more closely with the standard answer by tracing the transition from ancient methods to modern precision, even mentioning cultural influences on timekeeping.\\

            Overall, while both responses provide historical context, Response 2 is closer to the reference answer as it better follows the progression of timekeeping from ancient observations leading to the precise systems we have today.\\
            
            Therefore, \textcolor{magenta}{\textbf{Response 2}} is better. \textit{\textcolor[RGB]{202,12,22}{(The evaluation is correct!)}} \\
        \bottomrule
    \end{tabular}}
    \label{tab:success_2}
\end{table*}

%% file: tables/llm_question.tex
\begin{table*}[!htbp]
\centering
\caption{Prompt used for LLM-based question filtering.}
\begin{tabular}{p{0.95\textwidth}}
\toprule
\textbf{\textsc{Prompt for LLM Filter}} \\
\midrule
\textbf{Question Filtering Instructions} \\[2mm]

\textbf{Objective} \\
Filter out questions that are either unclear in their description or too broad to provide a meaningful reference. \\[2mm]

\textbf{Filtering Criteria} \\[1mm]
\textbf{1. Unclear Questions} \\
Reject questions that exhibit: \\
\quad * Ambiguous wording or phrasing \\
\quad * Multiple possible interpretations \\
\quad * Missing critical context or parameters \\
\quad * Vague or undefined terms \\
\quad * Grammatical issues that obscure meaning \\
\quad * Incomplete or fragmented thoughts \\[2mm]

\textbf{2. Overly Broad Questions} \\
Reject questions that: \\
\quad * Request information on topics with no reasonable boundaries \\
\quad * Would require encyclopedic or book-length answers \\
\quad * Ask for opinions on vast, multi-faceted subjects \\
\quad * Lack specific focus or scope constraints \\
\quad * Would yield references too general to be useful \\
\quad * Cover multiple unrelated topics simultaneously \\[2mm]

\textbf{Process} \\
1. Read the question carefully and completely \\
2. Evaluate against both clarity and breadth criteria \\
3. Make a filtering decision: \\
\quad \textbf{PASS}: Question is clear and appropriately scoped \\
\quad \textbf{REJECT - UNCLEAR}: Question lacks clarity (provide specific reason) \\
\quad \textbf{REJECT - TOO BROAD}: Question is overly broad (provide specific reason) \\[2mm]

\textbf{Examples of Questions to Reject} \\
\quad * ``What about technology?" (unclear) \\
\quad * ``Explain everything about human history" (too broad) \\
\quad * ``How does stuff work in general?" (both unclear and too broad) \\
\quad * ``What are all the factors affecting everything in the world?" (too broad) \\[2mm]

\textbf{Examples of Questions to Pass} \\
\quad * ``What is the boiling point of water at sea level?" \\
\quad * ``How does photosynthesis work in green plants?" \\
\quad * ``What were the main causes of World War I?" \\
\bottomrule
\end{tabular}
\label{tab:llm-filter}
\end{table*}

%% file: tables/llm_paraphrase.tex
\begin{table}[!htbp]
\centering
\caption{Prompt for LLM Paraphrase.}
\begin{tabular}{p{.9\textwidth}}
\toprule
\textbf{\textsc{Prompt for LLM Paraphrase}} \\
\midrule
\textbf{Objective:} \\
Transform the provided response into a more verbose version while strictly preserving the original meaning and information. \\[2mm]
\textbf{Requirements:} \\
- Expand the original text by adding descriptive language, elaborations, and explanatory phrases \\
- Maintain complete fidelity to the original information—do not introduce any new facts, claims, or insights \\
- Preserve the tone and intent of the original message \\
- Use stylistic techniques such as: \\
  \quad * Adding clarifying phrases and parenthetical explanations \\
  \quad * Employing more elaborate sentence structures \\
  \quad * Incorporating synonyms and varied vocabulary \\
  \quad * Adding transitional phrases between ideas \\
  \quad * Expanding brief points into full explanations \\
- Ensure the final text feels natural and not artificially inflated \\[2mm]
\textbf{Process:} \\
1. Thoroughly analyze the original response to understand its complete meaning \\
2. Identify core points and supporting details \\
3. Expand each point methodically while maintaining the original structure \\
4. Review to confirm no new information has been introduced \\
5. Polish the text for readability and flow \\
\bottomrule
\end{tabular}
\label{tab:english-llm-paraphrase}
\end{table}

%% file: tables/llm_evaluate_en.tex
\begin{table}[!htbp]
\centering
\caption{Prompt for English LLM Evaluation.}
\begin{tabular}{p{.9\textwidth}}
\toprule
\textbf{\textsc{Prompt for English LLM Evaluation}} \\
\midrule
We have the following question: \\

Question: {question} \\
The reference (standard) answer to this question is as follows: {reference} \\

We now have two student responses to this question. \\
Response 1 is as follows: {resp1} \\
Response 2 is as follows: {resp2} \\

Now, you should evaluate the two responses based on the content of the question, using the standard answer as the sole basis for judgment.\\ 
Determine which of the two—Response 1 or Response 2—is closer to the reference answer in terms of content. \\

Please begin with a brief analysis, and then provide your final judgment in one of the following forms: \\

If one response is better: \\
"Therefore, [Response 1] is better." or "Therefore, [Response 2] is better." \\

If the two responses are roughly equal in quality: \\
"Therefore, [Both responses are equal]." \\
\bottomrule
\end{tabular}
\label{tab:english-llm-eval}
\end{table}

%% file: tables/llm_pre.tex
\begin{table}[!htbp]
\centering
\caption{Prompt for Prometheus Evaluation.}
\begin{tabular}{p{.9\textwidth}}
\toprule
\textbf{\textsc{Prompt for Prometheus Evaluation}} \\
\midrule
\textbf{Task Description:} \\
An instruction (might include an Input inside it), a response to evaluate, and a score rubric representing a evaluation criteria are given. \\
1. Write a detailed feedback that assess the quality of two responses strictly based on the given score rubric, not evaluating in general. \\
2. After writing a feedback, choose a better response between Response A and Response B. You should refer to the score rubric. \\
3. The output format should look as follows: "Feedback: (write a feedback for criteria) [RESULT] (A or B)" \\
4. Please do not generate any other opening, closing, and explanations. \\

\textbf{Instruction:} \{orig\_instruction\} \\

\textbf{Response A:} \{response\_A\} \\

\textbf{Response B:} \{response\_B\} \\

\textbf{Score Rubric:} \{score\_rubric\} \\

\textbf{Feedback:} \\
\bottomrule
\end{tabular}
\label{tab:english-pro-eval}
\end{table}

%% file: tables/llm_autoj.tex
\begin{table}[!htbp]
\centering
\caption{Prompt for Auto-J Evaluation.}
\begin{tabular}{p{.9\textwidth}}
\toprule
\textbf{\textsc{Prompt for Auto-J Evaluation}} \\
\midrule
You are a helpful and precise assistant for checking the quality of the feedback. \\[2mm]
Two pieces of feedback have been provided for the same response to a particular query. Which one is better with regard to their correctness, comprehensiveness, and specificity to the query? \\[2mm]
\textbf{[BEGIN DATA]} \\

[Query]: \{prompt\} \\

[Response]: \{response\} \\

[Feedback 1]: \{feedback1\} \\

[Feedback 2]: \{feedback2\} \\

\textbf{[END DATA]} \\[2mm]
Please choose from the following options, and give out your reason in the next line. \\
A: Feedback 1 is significantly better. \\
B: Feedback 2 is significantly better. \\
C: Neither is significantly better. \\
\bottomrule
\end{tabular}
\label{tab:inst-autoj}
\end{table}

%% file: tables/annot_ref.tex
\begin{table*}[h]
\small
\centering
\caption{Annotation recipe for deciding valid references of \textsc{LFQA-E}.}
\begin{tabular}{p{.95\textwidth}}
\toprule
\textbf{\textsc{Annotation Recipe for Dropping Invalid References}} \\
\midrule
\textbf{Goal} \\
Keep only real, helpful explanation. Discard those that are uninformative, incorrect, or not actual explanations. \\[2mm]

\textbf{Keep if the answer:} \\
- Directly answers the question. \\
- Gives a simple but meaningful explanation (even if simplified). \\
- Is factually reasonable — not misleading or false. \\
- Stands alone (no ``see link" or ``I don't know"). \\[2mm]

\textbf{Good example:} "We yawn to help cool the brain and stay alert. It brings in oxygen and improves blood flow." \\[2mm]

\textbf{Discard if the answer is:} \\
- \textbf{Not an explanation} – e.g., ``Google it," ``It's magic," jokes, memes, one-liners (``Because science"). \\
- \textbf{Irrelevant} – Doesn't address the question or misunderstands it. \\
- \textbf{Factually wrong} – Clear misinformation (e.g., ``Rain comes from clouds crying"). \\
- \textbf{Too vague} – No real content: ``It's complicated," ``There are many reasons." \\
- \textbf{Avoids answering} – ``Great question!", ``Not sure, but here's a thought..." with nothing useful. \\
- \textbf{Circular} – Repeats the question: ``We sleep because we're tired." \\
- \textbf{Inappropriate} – Offensive, harmful, or unprofessional content. \\
\bottomrule
\end{tabular}
\label{tab:annot_ref}
\end{table*}

%% file: tables/annot.tex
\begin{table}[h]
\small
\centering
\caption{Annotation recipe of \textsc{LFQA-E}.}
\begin{tabular}{p{.95\textwidth}}
\toprule
\textbf{Overview} \\
This guide helps annotators evaluate and compare long-form responses against a reference to determine which response is more informative and complete. The process uses a triple-choice format (Response A Better, Response B Better, or Tie). \\[2mm]

\textbf{Key Principles} \\
- Focus on \textbf{factuality} and \textbf{completeness} according to the reference \\
- Fluency is not a primary evaluation criterion (all responses are expected to be fluent) \\
- Use information units as the basic evaluation unit \\
- Minimize bias through systematic comparison \\[2mm]

\textbf{Prerequisites} \\
- Domain knowledge relevant to the question topic \\
- Understanding of the subject matter through academic coursework or professional experience \\
- Ability to maintain focus during paragraph-level analysis \\[2mm]

\textbf{Evaluation Process} \\[1mm]
\textbf{Step 1: Extract Key Information from Reference} \\
1. Read the question carefully to understand what information is being requested \\
2. Read the reference thoroughly \\
3. Identify and list all key information units that: \\
   \quad - Directly answer the question \\
   \quad - Provide necessary context or background \\
   \quad - Support the main answer with evidence or examples \\
4. Organize key information into logical categories or themes \\[2mm]

\textbf{Step 2: Check for Key Information in Responses} \\
For each response (A and B): \\
1. Read the response completely \\
2. Map each key information unit from the reference to the response \\
3. Mark which key information units are: \\
   \quad - Present and accurate \\
   \quad - Present but inaccurate \\
   \quad - Missing entirely \\
4. Note any additional information not in the reference \\[2mm]

\textbf{Step 3: Handle Response Content} \\
1. Evaluate additional information: \\
   \quad - Is it relevant to the central topic? \\
   \quad - Does it enhance understanding or is it verbose/unnecessary? \\
2. Identify intertwined information: \\
   \quad - For sentences containing both correct and incorrect information, separate the components \\
   \quad - Assess the impact of any inaccuracies on the overall response quality \\[2mm]

\textbf{Step 4: Compare Overlapping Information} \\
1. Compare how well each response covers the key information units \\
2. Consider: \\
   \quad - Completeness: Which response includes more key information? \\
   \quad - Accuracy: Which response presents information more correctly? \\
   \quad - Relevance: Which response stays more focused on the question? \\
3. Compare the quality of overlapping information presentation \\[2mm]

\textbf{Step 5: Make Final Decision} \\
Select one of three options: \\
- \textbf{Response A is Better}: A contains more key information and/or presents it more accurately \\
- \textbf{Response B is Better}: B contains more key information and/or presents it more accurately \\
- \textbf{Tie}: Both responses are comparable in information coverage and accuracy \\[2mm]

\textbf{Common Pitfalls to Avoid} \\
1. Losing focus due to long paragraphs - use the systematic approach \\
2. Allowing domain bias to influence decisions - stick to the reference \\
3. Confusing eloquence with accuracy \\
4. Missing subtle differences between comparable responses \\
\bottomrule
\end{tabular}
\label{tab:annot}
\end{table}

%% file: iclr2026_conference.bib
@misc{fan2019eli5longformquestion,
      title={ELI5: Long Form Question Answering}, 
      author={Angela Fan and Yacine Jernite and Ethan Perez and David Grangier and Jason Weston and Michael Auli},
      year={2019},
      eprint={1907.09190},
      archivePrefix={arXiv},
      primaryClass={cs.CL},
      url={https://arxiv.org/abs/1907.09190}, 
}

@misc{rajpurkar2016squad100000questionsmachine,
      title={SQuAD: 100,000+ Questions for Machine Comprehension of Text}, 
      author={Pranav Rajpurkar and Jian Zhang and Konstantin Lopyrev and Percy Liang},
      year={2016},
      eprint={1606.05250},
      archivePrefix={arXiv},
      primaryClass={cs.CL},
      url={https://arxiv.org/abs/1606.05250}, 
}

@misc{joshi2017triviaqalargescaledistantly,
      title={TriviaQA: A Large Scale Distantly Supervised Challenge Dataset for Reading Comprehension}, 
      author={Mandar Joshi and Eunsol Choi and Daniel S. Weld and Luke Zettlemoyer},
      year={2017},
      eprint={1705.03551},
      archivePrefix={arXiv},
      primaryClass={cs.CL},
      url={https://arxiv.org/abs/1705.03551}, 
}

@misc{kočiský2017narrativeqareadingcomprehensionchallenge,
      title={The NarrativeQA Reading Comprehension Challenge}, 
      author={Tomáš Kočiský and Jonathan Schwarz and Phil Blunsom and Chris Dyer and Karl Moritz Hermann and Gábor Melis and Edward Grefenstette},
      year={2017},
      eprint={1712.07040},
      archivePrefix={arXiv},
      primaryClass={cs.CL},
      url={https://arxiv.org/abs/1712.07040}, 
}

@misc{xu2022answercomplexquestionsdiscourse,
      title={How Do We Answer Complex Questions: Discourse Structure of Long-form Answers}, 
      author={Fangyuan Xu and Junyi Jessy Li and Eunsol Choi},
      year={2022},
      eprint={2203.11048},
      archivePrefix={arXiv},
      primaryClass={cs.CL},
      url={https://arxiv.org/abs/2203.11048}, 
}

@misc{chen2023understandingretrievalaugmentationlongform,
      title={Understanding Retrieval Augmentation for Long-Form Question Answering}, 
      author={Hung-Ting Chen and Fangyuan Xu and Shane Arora and Eunsol Choi},
      year={2023},
      eprint={2310.12150},
      archivePrefix={arXiv},
      primaryClass={cs.CL},
      url={https://arxiv.org/abs/2310.12150}, 
}

@misc{akash2023longformquestionansweringiterative,
      title={Long-form Question Answering: An Iterative Planning-Retrieval-Generation Approach}, 
      author={Pritom Saha Akash and Kashob Kumar Roy and Lucian Popa and Kevin Chen-Chuan Chang},
      year={2023},
      eprint={2311.09383},
      archivePrefix={arXiv},
      primaryClass={cs.CL},
      url={https://arxiv.org/abs/2311.09383}, 
}

@inproceedings{lin-2004-rouge,
    title = "{ROUGE}: A Package for Automatic Evaluation of Summaries",
    author = "Lin, Chin-Yew",
    booktitle = "Text Summarization Branches Out",
    month = jul,
    year = "2004",
    address = "Barcelona, Spain",
    publisher = "Association for Computational Linguistics",
    url = "https://aclanthology.org/W04-1013",
    pages = "74--81",
}

@misc{krishna2021hurdlesprogresslongformquestion,
      title={Hurdles to Progress in Long-form Question Answering}, 
      author={Kalpesh Krishna and Aurko Roy and Mohit Iyyer},
      year={2021},
      eprint={2103.06332},
      archivePrefix={arXiv},
      primaryClass={cs.CL},
      url={https://arxiv.org/abs/2103.06332}, 
}

@misc{xu2023criticalevaluationevaluationslongform,
      title={A Critical Evaluation of Evaluations for Long-form Question Answering}, 
      author={Fangyuan Xu and Yixiao Song and Mohit Iyyer and Eunsol Choi},
      year={2023},
      eprint={2305.18201},
      archivePrefix={arXiv},
      primaryClass={cs.CL},
      url={https://arxiv.org/abs/2305.18201}, 
}

@misc{zhang2020bertscoreevaluatingtextgeneration,
      title={BERTScore: Evaluating Text Generation with BERT}, 
      author={Tianyi Zhang and Varsha Kishore and Felix Wu and Kilian Q. Weinberger and Yoav Artzi},
      year={2020},
      eprint={1904.09675},
      archivePrefix={arXiv},
      primaryClass={cs.CL},
      url={https://arxiv.org/abs/1904.09675}, 
}

@misc{yuan2021bartscoreevaluatinggeneratedtext,
      title={BARTScore: Evaluating Generated Text as Text Generation}, 
      author={Weizhe Yuan and Graham Neubig and Pengfei Liu},
      year={2021},
      eprint={2106.11520},
      archivePrefix={arXiv},
      primaryClass={cs.CL},
      url={https://arxiv.org/abs/2106.11520}, 
}

@misc{nakano2022webgptbrowserassistedquestionansweringhuman,
      title={WebGPT: Browser-assisted question-answering with human feedback}, 
      author={Reiichiro Nakano and Jacob Hilton and Suchir Balaji and Jeff Wu and Long Ouyang and Christina Kim and Christopher Hesse and Shantanu Jain and Vineet Kosaraju and William Saunders and Xu Jiang and Karl Cobbe and Tyna Eloundou and Gretchen Krueger and Kevin Button and Matthew Knight and Benjamin Chess and John Schulman},
      year={2022},
      eprint={2112.09332},
      archivePrefix={arXiv},
      primaryClass={cs.CL},
      url={https://arxiv.org/abs/2112.09332}, 
}

@misc{openai2024hello,
  title = {Hello GPT-4o},
  author = {{OpenAI}},
  year = {2024},
  howpublished = {\url{https://openai.com/index/hello-gpt-4o/}},
  note = {Accessed: 2024-08-05}
}

@misc{openai2023chatgpt,
  title = {ChatGPT: Chat Generative Pre-trained Transformer},
  author = {{OpenAI}},
  year = {2023},
  howpublished = {\url{https://chat.openai.com/}},
  note = {Accessed: 2024-08-05}
}

@misc{dubey2024llama3herdmodels,
      title={The Llama 3 Herd of Models}, 
      author={Abhimanyu Dubey and Abhinav Jauhri and Abhinav Pandey and Abhishek Kadian and Ahmad Al-Dahle and Aiesha Letman and Akhil Mathur and Alan Schelten and Amy Yang and Angela Fan and Anirudh Goyal and Anthony Hartshorn and Aobo Yang and Archi Mitra and Archie Sravankumar and Artem Korenev and Arthur Hinsvark and Arun Rao and Aston Zhang and Aurelien Rodriguez and Austen Gregerson and Ava Spataru and Baptiste Roziere and Bethany Biron and Binh Tang and Bobbie Chern and Charlotte Caucheteux and Chaya Nayak and Chloe Bi and Chris Marra and Chris McConnell and Christian Keller and Christophe Touret and Chunyang Wu and Corinne Wong and Cristian Canton Ferrer and Cyrus Nikolaidis and Damien Allonsius and Daniel Song and Danielle Pintz and Danny Livshits and David Esiobu and Dhruv Choudhary and Dhruv Mahajan and Diego Garcia-Olano and Diego Perino and Dieuwke Hupkes and Egor Lakomkin and Ehab AlBadawy and Elina Lobanova and Emily Dinan and Eric Michael Smith and Filip Radenovic and Frank Zhang and Gabriel Synnaeve and Gabrielle Lee and Georgia Lewis Anderson and Graeme Nail and Gregoire Mialon and Guan Pang and Guillem Cucurell and Hailey Nguyen and Hannah Korevaar and Hu Xu and Hugo Touvron and Iliyan Zarov and Imanol Arrieta Ibarra and Isabel Kloumann and Ishan Misra and Ivan Evtimov and Jade Copet and Jaewon Lee and Jan Geffert and Jana Vranes and Jason Park and Jay Mahadeokar and Jeet Shah and Jelmer van der Linde and Jennifer Billock and Jenny Hong and Jenya Lee and Jeremy Fu and Jianfeng Chi and Jianyu Huang and Jiawen Liu and Jie Wang and Jiecao Yu and Joanna Bitton and Joe Spisak and Jongsoo Park and Joseph Rocca and Joshua Johnstun and Joshua Saxe and Junteng Jia and Kalyan Vasuden Alwala and Kartikeya Upasani and Kate Plawiak and Ke Li and Kenneth Heafield and Kevin Stone and Khalid El-Arini and Krithika Iyer and Kshitiz Malik and Kuenley Chiu and Kunal Bhalla and Lauren Rantala-Yeary and Laurens van der Maaten and Lawrence Chen and Liang Tan and Liz Jenkins and Louis Martin and Lovish Madaan and Lubo Malo and Lukas Blecher and Lukas Landzaat and Luke de Oliveira and Madeline Muzzi and Mahesh Pasupuleti and Mannat Singh and Manohar Paluri and Marcin Kardas and Mathew Oldham and Mathieu Rita and Maya Pavlova and Melanie Kambadur and Mike Lewis and Min Si and Mitesh Kumar Singh and Mona Hassan and Naman Goyal and Narjes Torabi and Nikolay Bashlykov and Nikolay Bogoychev and Niladri Chatterji and Olivier Duchenne and Onur Çelebi and Patrick Alrassy and Pengchuan Zhang and Pengwei Li and Petar Vasic and Peter Weng and Prajjwal Bhargava and Pratik Dubal and Praveen Krishnan and Punit Singh Koura and Puxin Xu and Qing He and Qingxiao Dong and Ragavan Srinivasan and Raj Ganapathy and Ramon Calderer and Ricardo Silveira Cabral and Robert Stojnic and Roberta Raileanu and Rohit Girdhar and Rohit Patel and Romain Sauvestre and Ronnie Polidoro and Roshan Sumbaly and Ross Taylor and Ruan Silva and Rui Hou and Rui Wang and Saghar Hosseini and Sahana Chennabasappa and Sanjay Singh and Sean Bell and Seohyun Sonia Kim and Sergey Edunov and Shaoliang Nie and Sharan Narang and Sharath Raparthy and Sheng Shen and Shengye Wan and Shruti Bhosale and Shun Zhang and Simon Vandenhende and Soumya Batra and Spencer Whitman and Sten Sootla and Stephane Collot and Suchin Gururangan and Sydney Borodinsky and Tamar Herman and Tara Fowler and Tarek Sheasha and Thomas Georgiou and Thomas Scialom and Tobias Speckbacher and Todor Mihaylov and Tong Xiao and Ujjwal Karn and Vedanuj Goswami and Vibhor Gupta and Vignesh Ramanathan and Viktor Kerkez and Vincent Gonguet and Virginie Do and Vish Vogeti and Vladan Petrovic and Weiwei Chu and Wenhan Xiong and Wenyin Fu and Whitney Meers and Xavier Martinet and Xiaodong Wang and Xiaoqing Ellen Tan and Xinfeng Xie and Xuchao Jia and Xuewei Wang and Yaelle Goldschlag and Yashesh Gaur and Yasmine Babaei and Yi Wen and Yiwen Song and Yuchen Zhang and Yue Li and Yuning Mao and Zacharie Delpierre Coudert and Zheng Yan and Zhengxing Chen and Zoe Papakipos and Aaditya Singh and Aaron Grattafiori and Abha Jain and Adam Kelsey and Adam Shajnfeld and Adithya Gangidi and Adolfo Victoria and Ahuva Goldstand and Ajay Menon and Ajay Sharma and Alex Boesenberg and Alex Vaughan and Alexei Baevski and Allie Feinstein and Amanda Kallet and Amit Sangani and Anam Yunus and Andrei Lupu and Andres Alvarado and Andrew Caples and Andrew Gu and Andrew Ho and Andrew Poulton and Andrew Ryan and Ankit Ramchandani and Annie Franco and Aparajita Saraf and Arkabandhu Chowdhury and Ashley Gabriel and Ashwin Bharambe and Assaf Eisenman and Azadeh Yazdan and Beau James and Ben Maurer and Benjamin Leonhardi and Bernie Huang and Beth Loyd and Beto De Paola and Bhargavi Paranjape and Bing Liu and Bo Wu and Boyu Ni and Braden Hancock and Bram Wasti and Brandon Spence and Brani Stojkovic and Brian Gamido and Britt Montalvo and Carl Parker and Carly Burton and Catalina Mejia and Changhan Wang and Changkyu Kim and Chao Zhou and Chester Hu and Ching-Hsiang Chu and Chris Cai and Chris Tindal and Christoph Feichtenhofer and Damon Civin and Dana Beaty and Daniel Kreymer and Daniel Li and Danny Wyatt and David Adkins and David Xu and Davide Testuggine and Delia David and Devi Parikh and Diana Liskovich and Didem Foss and Dingkang Wang and Duc Le and Dustin Holland and Edward Dowling and Eissa Jamil and Elaine Montgomery and Eleonora Presani and Emily Hahn and Emily Wood and Erik Brinkman and Esteban Arcaute and Evan Dunbar and Evan Smothers and Fei Sun and Felix Kreuk and Feng Tian and Firat Ozgenel and Francesco Caggioni and Francisco Guzmán and Frank Kanayet and Frank Seide and Gabriela Medina Florez and Gabriella Schwarz and Gada Badeer and Georgia Swee and Gil Halpern and Govind Thattai and Grant Herman and Grigory Sizov and Guangyi and Zhang and Guna Lakshminarayanan and Hamid Shojanazeri and Han Zou and Hannah Wang and Hanwen Zha and Haroun Habeeb and Harrison Rudolph and Helen Suk and Henry Aspegren and Hunter Goldman and Igor Molybog and Igor Tufanov and Irina-Elena Veliche and Itai Gat and Jake Weissman and James Geboski and James Kohli and Japhet Asher and Jean-Baptiste Gaya and Jeff Marcus and Jeff Tang and Jennifer Chan and Jenny Zhen and Jeremy Reizenstein and Jeremy Teboul and Jessica Zhong and Jian Jin and Jingyi Yang and Joe Cummings and Jon Carvill and Jon Shepard and Jonathan McPhie and Jonathan Torres and Josh Ginsburg and Junjie Wang and Kai Wu and Kam Hou U and Karan Saxena and Karthik Prasad and Kartikay Khandelwal and Katayoun Zand and Kathy Matosich and Kaushik Veeraraghavan and Kelly Michelena and Keqian Li and Kun Huang and Kunal Chawla and Kushal Lakhotia and Kyle Huang and Lailin Chen and Lakshya Garg and Lavender A and Leandro Silva and Lee Bell and Lei Zhang and Liangpeng Guo and Licheng Yu and Liron Moshkovich and Luca Wehrstedt and Madian Khabsa and Manav Avalani and Manish Bhatt and Maria Tsimpoukelli and Martynas Mankus and Matan Hasson and Matthew Lennie and Matthias Reso and Maxim Groshev and Maxim Naumov and Maya Lathi and Meghan Keneally and Michael L. Seltzer and Michal Valko and Michelle Restrepo and Mihir Patel and Mik Vyatskov and Mikayel Samvelyan and Mike Clark and Mike Macey and Mike Wang and Miquel Jubert Hermoso and Mo Metanat and Mohammad Rastegari and Munish Bansal and Nandhini Santhanam and Natascha Parks and Natasha White and Navyata Bawa and Nayan Singhal and Nick Egebo and Nicolas Usunier and Nikolay Pavlovich Laptev and Ning Dong and Ning Zhang and Norman Cheng and Oleg Chernoguz and Olivia Hart and Omkar Salpekar and Ozlem Kalinli and Parkin Kent and Parth Parekh and Paul Saab and Pavan Balaji and Pedro Rittner and Philip Bontrager and Pierre Roux and Piotr Dollar and Polina Zvyagina and Prashant Ratanchandani and Pritish Yuvraj and Qian Liang and Rachad Alao and Rachel Rodriguez and Rafi Ayub and Raghotham Murthy and Raghu Nayani and Rahul Mitra and Raymond Li and Rebekkah Hogan and Robin Battey and Rocky Wang and Rohan Maheswari and Russ Howes and Ruty Rinott and Sai Jayesh Bondu and Samyak Datta and Sara Chugh and Sara Hunt and Sargun Dhillon and Sasha Sidorov and Satadru Pan and Saurabh Verma and Seiji Yamamoto and Sharadh Ramaswamy and Shaun Lindsay and Shaun Lindsay and Sheng Feng and Shenghao Lin and Shengxin Cindy Zha and Shiva Shankar and Shuqiang Zhang and Shuqiang Zhang and Sinong Wang and Sneha Agarwal and Soji Sajuyigbe and Soumith Chintala and Stephanie Max and Stephen Chen and Steve Kehoe and Steve Satterfield and Sudarshan Govindaprasad and Sumit Gupta and Sungmin Cho and Sunny Virk and Suraj Subramanian and Sy Choudhury and Sydney Goldman and Tal Remez and Tamar Glaser and Tamara Best and Thilo Kohler and Thomas Robinson and Tianhe Li and Tianjun Zhang and Tim Matthews and Timothy Chou and Tzook Shaked and Varun Vontimitta and Victoria Ajayi and Victoria Montanez and Vijai Mohan and Vinay Satish Kumar and Vishal Mangla and Vlad Ionescu and Vlad Poenaru and Vlad Tiberiu Mihailescu and Vladimir Ivanov and Wei Li and Wenchen Wang and Wenwen Jiang and Wes Bouaziz and Will Constable and Xiaocheng Tang and Xiaofang Wang and Xiaojian Wu and Xiaolan Wang and Xide Xia and Xilun Wu and Xinbo Gao and Yanjun Chen and Ye Hu and Ye Jia and Ye Qi and Yenda Li and Yilin Zhang and Ying Zhang and Yossi Adi and Youngjin Nam and Yu and Wang and Yuchen Hao and Yundi Qian and Yuzi He and Zach Rait and Zachary DeVito and Zef Rosnbrick and Zhaoduo Wen and Zhenyu Yang and Zhiwei Zhao},
      year={2024},
      eprint={2407.21783},
      archivePrefix={arXiv},
      primaryClass={cs.AI},
      url={https://arxiv.org/abs/2407.21783}, 
}

@misc{li2023generativejudgeevaluatingalignment,
      title={Generative Judge for Evaluating Alignment}, 
      author={Junlong Li and Shichao Sun and Weizhe Yuan and Run-Ze Fan and Hai Zhao and Pengfei Liu},
      year={2023},
      eprint={2310.05470},
      archivePrefix={arXiv},
      primaryClass={cs.CL},
      url={https://arxiv.org/abs/2310.05470}, 
}

@misc{jiang2024tigerscorebuildingexplainablemetric,
      title={TIGERScore: Towards Building Explainable Metric for All Text Generation Tasks}, 
      author={Dongfu Jiang and Yishan Li and Ge Zhang and Wenhao Huang and Bill Yuchen Lin and Wenhu Chen},
      year={2024},
      eprint={2310.00752},
      archivePrefix={arXiv},
      primaryClass={cs.CL},
      url={https://arxiv.org/abs/2310.00752}, 
}

@misc{wei2023chainofthoughtpromptingelicitsreasoning,
      title={Chain-of-Thought Prompting Elicits Reasoning in Large Language Models}, 
      author={Jason Wei and Xuezhi Wang and Dale Schuurmans and Maarten Bosma and Brian Ichter and Fei Xia and Ed Chi and Quoc Le and Denny Zhou},
      year={2023},
      eprint={2201.11903},
      archivePrefix={arXiv},
      primaryClass={cs.CL},
      url={https://arxiv.org/abs/2201.11903}, 
}

@misc{liu2023gevalnlgevaluationusing,
      title={G-Eval: NLG Evaluation using GPT-4 with Better Human Alignment}, 
      author={Yang Liu and Dan Iter and Yichong Xu and Shuohang Wang and Ruochen Xu and Chenguang Zhu},
      year={2023},
      eprint={2303.16634},
      archivePrefix={arXiv},
      primaryClass={cs.CL},
      url={https://arxiv.org/abs/2303.16634}, 
}

@misc{chiang2024chatbot,
    title={Chatbot Arena: An Open Platform for Evaluating LLMs by Human Preference},
    author={Wei-Lin Chiang and Lianmin Zheng and Ying Sheng and Anastasios Nikolas Angelopoulos and Tianle Li and Dacheng Li and Hao Zhang and Banghua Zhu and Michael Jordan and Joseph E. Gonzalez and Ion Stoica},
    year={2024},
    eprint={2403.04132},
    archivePrefix={arXiv},
    primaryClass={cs.AI}
}

@inproceedings{zheng2023judging,
    title={Judging LLM-as-a-Judge with MT-Bench and Chatbot Arena},
    author={Lianmin Zheng and Wei-Lin Chiang and Ying Sheng and Siyuan Zhuang and Zhanghao Wu and Yonghao Zhuang and Zi Lin and Zhuohan Li and Dacheng Li and Eric Xing and Hao Zhang and Joseph E. Gonzalez and Ion Stoica},
    booktitle={Thirty-seventh Conference on Neural Information Processing Systems Datasets and Benchmarks Track},
    year={2023},
    url={https://openreview.net/forum?id=uccHPGDlao}
}

@inproceedings{zheng2024lmsyschatm,
    title={LMSYS-Chat-1M: A Large-Scale Real-World LLM Conversation Dataset},
    author={Lianmin Zheng and Wei-Lin Chiang and Ying Sheng and Tianle Li and Siyuan Zhuang and Zhanghao Wu and Yonghao Zhuang and Zhuohan Li and Zi Lin and Eric Xing and Joseph E. Gonzalez and Ion Stoica and Hao Zhang},
    booktitle={The Twelfth International Conference on Learning Representations},
    year={2024},
    url={https://openreview.net/forum?id=BOfDKxfwt0}
}

@article{fan2024eva,
  title={EVA-Score: Evaluating Abstractive Long-form Summarization on Informativeness through Extraction and Validation},
  author={Fan, Yuchen and Zhong, Xin and Wan, Yazhe and Wang, Chengsi and Cheng, Haonan and Wu, Gaoche and Ding, Ning and Zhou, Bowen},
  journal={arXiv preprint arXiv:2407.04969},
  year={2024}
}

@article{fan2024evaluating,
  title={Evaluating generative language models in information extraction as subjective question correction},
  author={Fan, Yuchen and Liu, Yantao and Yao, Zijun and Yu, Jifan and Hou, Lei and Li, Juanzi},
  journal={arXiv preprint arXiv:2404.03532},
  year={2024}
}

@misc{qwen2025qwen25technicalreport,
      title={Qwen2.5 Technical Report}, 
      author={Qwen and : and An Yang and Baosong Yang and Beichen Zhang and Binyuan Hui and Bo Zheng and Bowen Yu and Chengyuan Li and Dayiheng Liu and Fei Huang and Haoran Wei and Huan Lin and Jian Yang and Jianhong Tu and Jianwei Zhang and Jianxin Yang and Jiaxi Yang and Jingren Zhou and Junyang Lin and Kai Dang and Keming Lu and Keqin Bao and Kexin Yang and Le Yu and Mei Li and Mingfeng Xue and Pei Zhang and Qin Zhu and Rui Men and Runji Lin and Tianhao Li and Tianyi Tang and Tingyu Xia and Xingzhang Ren and Xuancheng Ren and Yang Fan and Yang Su and Yichang Zhang and Yu Wan and Yuqiong Liu and Zeyu Cui and Zhenru Zhang and Zihan Qiu},
      year={2025},
      eprint={2412.15115},
      archivePrefix={arXiv},
      primaryClass={cs.CL},
      url={https://arxiv.org/abs/2412.15115}, 
}

@misc{deepseekai2025deepseekv3technicalreport,
      title={DeepSeek-V3 Technical Report}, 
      author={DeepSeek-AI and Aixin Liu and Bei Feng and Bing Xue and Bingxuan Wang and Bochao Wu and Chengda Lu and Chenggang Zhao and Chengqi Deng and Chenyu Zhang and Chong Ruan and Damai Dai and Daya Guo and Dejian Yang and Deli Chen and Dongjie Ji and Erhang Li and Fangyun Lin and Fucong Dai and Fuli Luo and Guangbo Hao and Guanting Chen and Guowei Li and H. Zhang and Han Bao and Hanwei Xu and Haocheng Wang and Haowei Zhang and Honghui Ding and Huajian Xin and Huazuo Gao and Hui Li and Hui Qu and J. L. Cai and Jian Liang and Jianzhong Guo and Jiaqi Ni and Jiashi Li and Jiawei Wang and Jin Chen and Jingchang Chen and Jingyang Yuan and Junjie Qiu and Junlong Li and Junxiao Song and Kai Dong and Kai Hu and Kaige Gao and Kang Guan and Kexin Huang and Kuai Yu and Lean Wang and Lecong Zhang and Lei Xu and Leyi Xia and Liang Zhao and Litong Wang and Liyue Zhang and Meng Li and Miaojun Wang and Mingchuan Zhang and Minghua Zhang and Minghui Tang and Mingming Li and Ning Tian and Panpan Huang and Peiyi Wang and Peng Zhang and Qiancheng Wang and Qihao Zhu and Qinyu Chen and Qiushi Du and R. J. Chen and R. L. Jin and Ruiqi Ge and Ruisong Zhang and Ruizhe Pan and Runji Wang and Runxin Xu and Ruoyu Zhang and Ruyi Chen and S. S. Li and Shanghao Lu and Shangyan Zhou and Shanhuang Chen and Shaoqing Wu and Shengfeng Ye and Shengfeng Ye and Shirong Ma and Shiyu Wang and Shuang Zhou and Shuiping Yu and Shunfeng Zhou and Shuting Pan and T. Wang and Tao Yun and Tian Pei and Tianyu Sun and W. L. Xiao and Wangding Zeng and Wanjia Zhao and Wei An and Wen Liu and Wenfeng Liang and Wenjun Gao and Wenqin Yu and Wentao Zhang and X. Q. Li and Xiangyue Jin and Xianzu Wang and Xiao Bi and Xiaodong Liu and Xiaohan Wang and Xiaojin Shen and Xiaokang Chen and Xiaokang Zhang and Xiaosha Chen and Xiaotao Nie and Xiaowen Sun and Xiaoxiang Wang and Xin Cheng and Xin Liu and Xin Xie and Xingchao Liu and Xingkai Yu and Xinnan Song and Xinxia Shan and Xinyi Zhou and Xinyu Yang and Xinyuan Li and Xuecheng Su and Xuheng Lin and Y. K. Li and Y. Q. Wang and Y. X. Wei and Y. X. Zhu and Yang Zhang and Yanhong Xu and Yanhong Xu and Yanping Huang and Yao Li and Yao Zhao and Yaofeng Sun and Yaohui Li and Yaohui Wang and Yi Yu and Yi Zheng and Yichao Zhang and Yifan Shi and Yiliang Xiong and Ying He and Ying Tang and Yishi Piao and Yisong Wang and Yixuan Tan and Yiyang Ma and Yiyuan Liu and Yongqiang Guo and Yu Wu and Yuan Ou and Yuchen Zhu and Yuduan Wang and Yue Gong and Yuheng Zou and Yujia He and Yukun Zha and Yunfan Xiong and Yunxian Ma and Yuting Yan and Yuxiang Luo and Yuxiang You and Yuxuan Liu and Yuyang Zhou and Z. F. Wu and Z. Z. Ren and Zehui Ren and Zhangli Sha and Zhe Fu and Zhean Xu and Zhen Huang and Zhen Zhang and Zhenda Xie and Zhengyan Zhang and Zhewen Hao and Zhibin Gou and Zhicheng Ma and Zhigang Yan and Zhihong Shao and Zhipeng Xu and Zhiyu Wu and Zhongyu Zhang and Zhuoshu Li and Zihui Gu and Zijia Zhu and Zijun Liu and Zilin Li and Ziwei Xie and Ziyang Song and Ziyi Gao and Zizheng Pan},
      year={2025},
      eprint={2412.19437},
      archivePrefix={arXiv},
      primaryClass={cs.CL},
      url={https://arxiv.org/abs/2412.19437}, 
}

@misc{openai2024gpt4ocard,
      title={GPT-4o System Card}, 
      author={OpenAI and : and Aaron Hurst and Adam Lerer and Adam P. Goucher and Adam Perelman and Aditya Ramesh and Aidan Clark and AJ Ostrow and Akila Welihinda and Alan Hayes and Alec Radford and Aleksander Mądry and Alex Baker-Whitcomb and Alex Beutel and Alex Borzunov and Alex Carney and Alex Chow and Alex Kirillov and Alex Nichol and Alex Paino and Alex Renzin and Alex Tachard Passos and Alexander Kirillov and Alexi Christakis and Alexis Conneau and Ali Kamali and Allan Jabri and Allison Moyer and Allison Tam and Amadou Crookes and Amin Tootoochian and Amin Tootoonchian and Ananya Kumar and Andrea Vallone and Andrej Karpathy and Andrew Braunstein and Andrew Cann and Andrew Codispoti and Andrew Galu and Andrew Kondrich and Andrew Tulloch and Andrey Mishchenko and Angela Baek and Angela Jiang and Antoine Pelisse and Antonia Woodford and Anuj Gosalia and Arka Dhar and Ashley Pantuliano and Avi Nayak and Avital Oliver and Barret Zoph and Behrooz Ghorbani and Ben Leimberger and Ben Rossen and Ben Sokolowsky and Ben Wang and Benjamin Zweig and Beth Hoover and Blake Samic and Bob McGrew and Bobby Spero and Bogo Giertler and Bowen Cheng and Brad Lightcap and Brandon Walkin and Brendan Quinn and Brian Guarraci and Brian Hsu and Bright Kellogg and Brydon Eastman and Camillo Lugaresi and Carroll Wainwright and Cary Bassin and Cary Hudson and Casey Chu and Chad Nelson and Chak Li and Chan Jun Shern and Channing Conger and Charlotte Barette and Chelsea Voss and Chen Ding and Cheng Lu and Chong Zhang and Chris Beaumont and Chris Hallacy and Chris Koch and Christian Gibson and Christina Kim and Christine Choi and Christine McLeavey and Christopher Hesse and Claudia Fischer and Clemens Winter and Coley Czarnecki and Colin Jarvis and Colin Wei and Constantin Koumouzelis and Dane Sherburn and Daniel Kappler and Daniel Levin and Daniel Levy and David Carr and David Farhi and David Mely and David Robinson and David Sasaki and Denny Jin and Dev Valladares and Dimitris Tsipras and Doug Li and Duc Phong Nguyen and Duncan Findlay and Edede Oiwoh and Edmund Wong and Ehsan Asdar and Elizabeth Proehl and Elizabeth Yang and Eric Antonow and Eric Kramer and Eric Peterson and Eric Sigler and Eric Wallace and Eugene Brevdo and Evan Mays and Farzad Khorasani and Felipe Petroski Such and Filippo Raso and Francis Zhang and Fred von Lohmann and Freddie Sulit and Gabriel Goh and Gene Oden and Geoff Salmon and Giulio Starace and Greg Brockman and Hadi Salman and Haiming Bao and Haitang Hu and Hannah Wong and Haoyu Wang and Heather Schmidt and Heather Whitney and Heewoo Jun and Hendrik Kirchner and Henrique Ponde de Oliveira Pinto and Hongyu Ren and Huiwen Chang and Hyung Won Chung and Ian Kivlichan and Ian O'Connell and Ian O'Connell and Ian Osband and Ian Silber and Ian Sohl and Ibrahim Okuyucu and Ikai Lan and Ilya Kostrikov and Ilya Sutskever and Ingmar Kanitscheider and Ishaan Gulrajani and Jacob Coxon and Jacob Menick and Jakub Pachocki and James Aung and James Betker and James Crooks and James Lennon and Jamie Kiros and Jan Leike and Jane Park and Jason Kwon and Jason Phang and Jason Teplitz and Jason Wei and Jason Wolfe and Jay Chen and Jeff Harris and Jenia Varavva and Jessica Gan Lee and Jessica Shieh and Ji Lin and Jiahui Yu and Jiayi Weng and Jie Tang and Jieqi Yu and Joanne Jang and Joaquin Quinonero Candela and Joe Beutler and Joe Landers and Joel Parish and Johannes Heidecke and John Schulman and Jonathan Lachman and Jonathan McKay and Jonathan Uesato and Jonathan Ward and Jong Wook Kim and Joost Huizinga and Jordan Sitkin and Jos Kraaijeveld and Josh Gross and Josh Kaplan and Josh Snyder and Joshua Achiam and Joy Jiao and Joyce Lee and Juntang Zhuang and Justyn Harriman and Kai Fricke and Kai Hayashi and Karan Singhal and Katy Shi and Kavin Karthik and Kayla Wood and Kendra Rimbach and Kenny Hsu and Kenny Nguyen and Keren Gu-Lemberg and Kevin Button and Kevin Liu and Kiel Howe and Krithika Muthukumar and Kyle Luther and Lama Ahmad and Larry Kai and Lauren Itow and Lauren Workman and Leher Pathak and Leo Chen and Li Jing and Lia Guy and Liam Fedus and Liang Zhou and Lien Mamitsuka and Lilian Weng and Lindsay McCallum and Lindsey Held and Long Ouyang and Louis Feuvrier and Lu Zhang and Lukas Kondraciuk and Lukasz Kaiser and Luke Hewitt and Luke Metz and Lyric Doshi and Mada Aflak and Maddie Simens and Madelaine Boyd and Madeleine Thompson and Marat Dukhan and Mark Chen and Mark Gray and Mark Hudnall and Marvin Zhang and Marwan Aljubeh and Mateusz Litwin and Matthew Zeng and Max Johnson and Maya Shetty and Mayank Gupta and Meghan Shah and Mehmet Yatbaz and Meng Jia Yang and Mengchao Zhong and Mia Glaese and Mianna Chen and Michael Janner and Michael Lampe and Michael Petrov and Michael Wu and Michele Wang and Michelle Fradin and Michelle Pokrass and Miguel Castro and Miguel Oom Temudo de Castro and Mikhail Pavlov and Miles Brundage and Miles Wang and Minal Khan and Mira Murati and Mo Bavarian and Molly Lin and Murat Yesildal and Nacho Soto and Natalia Gimelshein and Natalie Cone and Natalie Staudacher and Natalie Summers and Natan LaFontaine and Neil Chowdhury and Nick Ryder and Nick Stathas and Nick Turley and Nik Tezak and Niko Felix and Nithanth Kudige and Nitish Keskar and Noah Deutsch and Noel Bundick and Nora Puckett and Ofir Nachum and Ola Okelola and Oleg Boiko and Oleg Murk and Oliver Jaffe and Olivia Watkins and Olivier Godement and Owen Campbell-Moore and Patrick Chao and Paul McMillan and Pavel Belov and Peng Su and Peter Bak and Peter Bakkum and Peter Deng and Peter Dolan and Peter Hoeschele and Peter Welinder and Phil Tillet and Philip Pronin and Philippe Tillet and Prafulla Dhariwal and Qiming Yuan and Rachel Dias and Rachel Lim and Rahul Arora and Rajan Troll and Randall Lin and Rapha Gontijo Lopes and Raul Puri and Reah Miyara and Reimar Leike and Renaud Gaubert and Reza Zamani and Ricky Wang and Rob Donnelly and Rob Honsby and Rocky Smith and Rohan Sahai and Rohit Ramchandani and Romain Huet and Rory Carmichael and Rowan Zellers and Roy Chen and Ruby Chen and Ruslan Nigmatullin and Ryan Cheu and Saachi Jain and Sam Altman and Sam Schoenholz and Sam Toizer and Samuel Miserendino and Sandhini Agarwal and Sara Culver and Scott Ethersmith and Scott Gray and Sean Grove and Sean Metzger and Shamez Hermani and Shantanu Jain and Shengjia Zhao and Sherwin Wu and Shino Jomoto and Shirong Wu and Shuaiqi and Xia and Sonia Phene and Spencer Papay and Srinivas Narayanan and Steve Coffey and Steve Lee and Stewart Hall and Suchir Balaji and Tal Broda and Tal Stramer and Tao Xu and Tarun Gogineni and Taya Christianson and Ted Sanders and Tejal Patwardhan and Thomas Cunninghman and Thomas Degry and Thomas Dimson and Thomas Raoux and Thomas Shadwell and Tianhao Zheng and Todd Underwood and Todor Markov and Toki Sherbakov and Tom Rubin and Tom Stasi and Tomer Kaftan and Tristan Heywood and Troy Peterson and Tyce Walters and Tyna Eloundou and Valerie Qi and Veit Moeller and Vinnie Monaco and Vishal Kuo and Vlad Fomenko and Wayne Chang and Weiyi Zheng and Wenda Zhou and Wesam Manassra and Will Sheu and Wojciech Zaremba and Yash Patil and Yilei Qian and Yongjik Kim and Youlong Cheng and Yu Zhang and Yuchen He and Yuchen Zhang and Yujia Jin and Yunxing Dai and Yury Malkov},
      year={2024},
      eprint={2410.21276},
      archivePrefix={arXiv},
      primaryClass={cs.CL},
      url={https://arxiv.org/abs/2410.21276}, 
}

@misc{deepseekai2025deepseekr1incentivizingreasoningcapability,
      title={DeepSeek-R1: Incentivizing Reasoning Capability in LLMs via Reinforcement Learning}, 
      author={DeepSeek-AI and Daya Guo and Dejian Yang and Haowei Zhang and Junxiao Song and Ruoyu Zhang and Runxin Xu and Qihao Zhu and Shirong Ma and Peiyi Wang and Xiao Bi and Xiaokang Zhang and Xingkai Yu and Yu Wu and Z. F. Wu and Zhibin Gou and Zhihong Shao and Zhuoshu Li and Ziyi Gao and Aixin Liu and Bing Xue and Bingxuan Wang and Bochao Wu and Bei Feng and Chengda Lu and Chenggang Zhao and Chengqi Deng and Chenyu Zhang and Chong Ruan and Damai Dai and Deli Chen and Dongjie Ji and Erhang Li and Fangyun Lin and Fucong Dai and Fuli Luo and Guangbo Hao and Guanting Chen and Guowei Li and H. Zhang and Han Bao and Hanwei Xu and Haocheng Wang and Honghui Ding and Huajian Xin and Huazuo Gao and Hui Qu and Hui Li and Jianzhong Guo and Jiashi Li and Jiawei Wang and Jingchang Chen and Jingyang Yuan and Junjie Qiu and Junlong Li and J. L. Cai and Jiaqi Ni and Jian Liang and Jin Chen and Kai Dong and Kai Hu and Kaige Gao and Kang Guan and Kexin Huang and Kuai Yu and Lean Wang and Lecong Zhang and Liang Zhao and Litong Wang and Liyue Zhang and Lei Xu and Leyi Xia and Mingchuan Zhang and Minghua Zhang and Minghui Tang and Meng Li and Miaojun Wang and Mingming Li and Ning Tian and Panpan Huang and Peng Zhang and Qiancheng Wang and Qinyu Chen and Qiushi Du and Ruiqi Ge and Ruisong Zhang and Ruizhe Pan and Runji Wang and R. J. Chen and R. L. Jin and Ruyi Chen and Shanghao Lu and Shangyan Zhou and Shanhuang Chen and Shengfeng Ye and Shiyu Wang and Shuiping Yu and Shunfeng Zhou and Shuting Pan and S. S. Li and Shuang Zhou and Shaoqing Wu and Shengfeng Ye and Tao Yun and Tian Pei and Tianyu Sun and T. Wang and Wangding Zeng and Wanjia Zhao and Wen Liu and Wenfeng Liang and Wenjun Gao and Wenqin Yu and Wentao Zhang and W. L. Xiao and Wei An and Xiaodong Liu and Xiaohan Wang and Xiaokang Chen and Xiaotao Nie and Xin Cheng and Xin Liu and Xin Xie and Xingchao Liu and Xinyu Yang and Xinyuan Li and Xuecheng Su and Xuheng Lin and X. Q. Li and Xiangyue Jin and Xiaojin Shen and Xiaosha Chen and Xiaowen Sun and Xiaoxiang Wang and Xinnan Song and Xinyi Zhou and Xianzu Wang and Xinxia Shan and Y. K. Li and Y. Q. Wang and Y. X. Wei and Yang Zhang and Yanhong Xu and Yao Li and Yao Zhao and Yaofeng Sun and Yaohui Wang and Yi Yu and Yichao Zhang and Yifan Shi and Yiliang Xiong and Ying He and Yishi Piao and Yisong Wang and Yixuan Tan and Yiyang Ma and Yiyuan Liu and Yongqiang Guo and Yuan Ou and Yuduan Wang and Yue Gong and Yuheng Zou and Yujia He and Yunfan Xiong and Yuxiang Luo and Yuxiang You and Yuxuan Liu and Yuyang Zhou and Y. X. Zhu and Yanhong Xu and Yanping Huang and Yaohui Li and Yi Zheng and Yuchen Zhu and Yunxian Ma and Ying Tang and Yukun Zha and Yuting Yan and Z. Z. Ren and Zehui Ren and Zhangli Sha and Zhe Fu and Zhean Xu and Zhenda Xie and Zhengyan Zhang and Zhewen Hao and Zhicheng Ma and Zhigang Yan and Zhiyu Wu and Zihui Gu and Zijia Zhu and Zijun Liu and Zilin Li and Ziwei Xie and Ziyang Song and Zizheng Pan and Zhen Huang and Zhipeng Xu and Zhongyu Zhang and Zhen Zhang},
      year={2025},
      eprint={2501.12948},
      archivePrefix={arXiv},
      primaryClass={cs.CL},
      url={https://arxiv.org/abs/2501.12948}, 
}

@article{liu2024skywork,
  title={Skywork-Reward: Bag of Tricks for Reward Modeling in LLMs},
  author={Liu, Chris Yuhao and Zeng, Liang and Liu, Jiacai and Yan, Rui and He, Jujie and Wang, Chaojie and Yan, Shuicheng and Liu, Yang and Zhou, Yahui},
  journal={arXiv preprint arXiv:2410.18451},
  year={2024}
}

@misc{RewardBench,
    title={RewardBench: Evaluating Reward Models for Language Modeling},
    author={Lambert, Nathan and Pyatkin, Valentina and Morrison, Jacob and Miranda, LJ and Lin, Bill Yuchen and Chandu, Khyathi and Dziri, Nouha and Kumar, Sachin and Zick, Tom and Choi, Yejin and Smith, Noah A. and Hajishirzi, Hannaneh},
    year={2024},
    howpublished={\url{https://huggingface.co/spaces/allenai/reward-bench}}
}

@misc{kim2024prometheus,
      title={Prometheus 2: An Open Source Language Model Specialized in Evaluating Other Language Models}, 
      author={Seungone Kim and Juyoung Suk and Shayne Longpre and Bill Yuchen Lin and Jamin Shin and Sean Welleck and Graham Neubig and Moontae Lee and Kyungjae Lee and Minjoon Seo},
      year={2024},
      eprint={2405.01535},
      archivePrefix={arXiv},
      primaryClass={cs.CL}
}

@misc{chen2025rmr1rewardmodelingreasoning,
      title={RM-R1: Reward Modeling as Reasoning}, 
      author={Xiusi Chen and Gaotang Li and Ziqi Wang and Bowen Jin and Cheng Qian and Yu Wang and Hongru Wang and Yu Zhang and Denghui Zhang and Tong Zhang and Hanghang Tong and Heng Ji},
      year={2025},
      eprint={2505.02387},
      archivePrefix={arXiv},
      primaryClass={cs.CL},
      url={https://arxiv.org/abs/2505.02387}, 
}

@misc{liu2024rmbenchbenchmarkingrewardmodels,
      title={RM-Bench: Benchmarking Reward Models of Language Models with Subtlety and Style}, 
      author={Yantao Liu and Zijun Yao and Rui Min and Yixin Cao and Lei Hou and Juanzi Li},
      year={2024},
      eprint={2410.16184},
      archivePrefix={arXiv},
      primaryClass={cs.CL},
      url={https://arxiv.org/abs/2410.16184}, 
}

@misc{min2023factscorefinegrainedatomicevaluation,
      title={FActScore: Fine-grained Atomic Evaluation of Factual Precision in Long Form Text Generation}, 
      author={Sewon Min and Kalpesh Krishna and Xinxi Lyu and Mike Lewis and Wen-tau Yih and Pang Wei Koh and Mohit Iyyer and Luke Zettlemoyer and Hannaneh Hajishirzi},
      year={2023},
      eprint={2305.14251},
      archivePrefix={arXiv},
      primaryClass={cs.CL},
      url={https://arxiv.org/abs/2305.14251}, 
}

@misc{chang2023surveyevaluationlargelanguage,
      title={A Survey on Evaluation of Large Language Models}, 
      author={Yupeng Chang and Xu Wang and Jindong Wang and Yuan Wu and Linyi Yang and Kaijie Zhu and Hao Chen and Xiaoyuan Yi and Cunxiang Wang and Yidong Wang and Wei Ye and Yue Zhang and Yi Chang and Philip S. Yu and Qiang Yang and Xing Xie},
      year={2023},
      eprint={2307.03109},
      archivePrefix={arXiv},
      primaryClass={cs.CL},
      url={https://arxiv.org/abs/2307.03109}, 
}

@misc{liu2024rmr1,
      title={RM-R1: Reward Modeling as Reasoning}, 
      author={Xiusi Liu and Gaotang Li and Ziqi Wang and Bowen Jin and Cheng Qian and Yu Wang and Hongru Wang and Yu Zhang and Denghui Zhang and Tong Zhang and Hanghang Tong and Heng Ji},
      year={2024},
      eprint={2505.02387},
      archivePrefix={arXiv},
      primaryClass={cs.CL},
      url={https://arxiv.org/abs/2505.02387},
}

@article{xu2024benchmarking,
      title={Benchmarking Benchmark Leakage in Large Language Models}, 
      author={Xu, Ruijie and Wang, Zengzhi and Fan, Run-Ze and Liu, Pengfei},
      year={2024},
      journal={arXiv preprint arXiv:2404.18824},
      url={https://arxiv.org/abs/2404.18824}
}

@article{fan2025ssrl,
  title={SSRL: Self-Search Reinforcement Learning},
  author={Fan, Yuchen and Zhang, Kaiyan and Zhou, Heng and Zuo, Yuxin and Chen, Yanxu and Fu, Yu and Long, Xinwei and Zhu, Xuekai and Jiang, Che and Zhang, Yuchen and others},
  journal={arXiv preprint arXiv:2508.10874},
  year={2025}
}

@article{cui2025process,
  title={Process reinforcement through implicit rewards},
  author={Cui, Ganqu and Yuan, Lifan and Wang, Zefan and Wang, Hanbin and Li, Wendi and He, Bingxiang and Fan, Yuchen and Yu, Tianyu and Xu, Qixin and Chen, Weize and others},
  journal={arXiv preprint arXiv:2502.01456},
  year={2025}
}

@article{zuo2025ttrl,
  title={Ttrl: Test-time reinforcement learning},
  author={Zuo, Yuxin and Zhang, Kaiyan and Sheng, Li and Qu, Shang and Cui, Ganqu and Zhu, Xuekai and Li, Haozhan and Zhang, Yuchen and Long, Xinwei and Hua, Ermo and others},
  journal={arXiv preprint arXiv:2504.16084},
  year={2025}
}

@misc{yu2025dapoopensourcellmreinforcement,
      title={DAPO: An Open-Source LLM Reinforcement Learning System at Scale}, 
      author={Qiying Yu and Zheng Zhang and Ruofei Zhu and Yufeng Yuan and Xiaochen Zuo and Yu Yue and Weinan Dai and Tiantian Fan and Gaohong Liu and Lingjun Liu and Xin Liu and Haibin Lin and Zhiqi Lin and Bole Ma and Guangming Sheng and Yuxuan Tong and Chi Zhang and Mofan Zhang and Wang Zhang and Hang Zhu and Jinhua Zhu and Jiaze Chen and Jiangjie Chen and Chengyi Wang and Hongli Yu and Yuxuan Song and Xiangpeng Wei and Hao Zhou and Jingjing Liu and Wei-Ying Ma and Ya-Qin Zhang and Lin Yan and Mu Qiao and Yonghui Wu and Mingxuan Wang},
      year={2025},
      eprint={2503.14476},
      archivePrefix={arXiv},
      primaryClass={cs.LG},
      url={https://arxiv.org/abs/2503.14476}, 
}

@article{zhou2025reso,
  title={Reso: A reward-driven self-organizing llm-based multi-agent system for reasoning tasks},
  author={Zhou, Heng and Geng, Hejia and Xue, Xiangyuan and Kang, Li and Qin, Yiran and Wang, Zhiyong and Yin, Zhenfei and Bai, Lei},
  journal={arXiv preprint arXiv:2503.02390},
  year={2025}
}

@article{zhang2025survey,
  title={A survey of reinforcement learning for large reasoning models},
  author={Zhang, Kaiyan and Zuo, Yuxin and He, Bingxiang and Sun, Youbang and Liu, Runze and Jiang, Che and Fan, Yuchen and Tian, Kai and Jia, Guoli and Li, Pengfei and others},
  journal={arXiv preprint arXiv:2509.08827},
  year={2025}
}
